\documentclass{article}

 \usepackage[main, final]{neurips_2026}

\usepackage[utf8]{inputenc} 
\usepackage[T1]{fontenc}    
\usepackage[hypertexnames=false]{hyperref}       
\usepackage{url}            
\usepackage{booktabs}       
\usepackage{graphicx} 

\usepackage{nicefrac}       
\usepackage{microtype}      
\usepackage[most]{tcolorbox}
\usepackage{xcolor}         
\usepackage{amsmath,amssymb,mathtools,amsthm,amsfonts,dsfont,bm, thmtools}
\usepackage{algorithm}
\usepackage{algorithmic}
\usepackage[normalem]{ulem}
\usepackage{array}
\usepackage{multirow, multicol}
\usepackage[capitalize,noabbrev]{cleveref}

\theoremstyle{plain}
\newtheorem{theorem}{Theorem}[section]
\newtheorem{proposition}{Proposition}
\theoremstyle{remark}
\newtheorem{remark}{Remark}

\makeatletter
\let\c@proposition\c@theorem
\let\c@remark\c@theorem

\makeatother

\usepackage[textsize=tiny]{todonotes}
\usepackage{float}

\newcommand\blfootnote[1]{%
  \begingroup
  \renewcommand\thefootnote{}\footnote{#1}%
  \addtocounter{footnote}{-1}%
  \endgroup
}
\newcommand{\nc}{\newcommand}
\nc{\nn}{\noindent}
\nc{\dmo}{\DeclareMathOperator}
\nc{\dc}{\definecolor}
\dc{watpink}{RGB}{198, 0, 120}
\dc{skyblue}{RGB}{0, 102, 204}
\dc{darkgreen}{RGB}{0, 100, 0}
\dc{gsublue}{RGB}{0, 57, 166}
\nc{\wpink}[1]{\textcolor{watpink}{#1}}
\nc{\yck}[1]{\wpink{[YCK: #1]}}
\nc{\wblue}[1]{\textcolor{blue}{#1}}
\nc{\ycky}[1]{\textcolor{gsublue}{[YCK: #1]}}

\nc{\lightblue}[1]{\textcolor{lightblue}{#1}}

\nc{\zj}[1]{%
  \todo[linecolor=blue,backgroundcolor=blue!25,bordercolor=blue]{Zijun: #1}%
}
\nc{\xy}[1]{%
  \todo[linecolor=red,backgroundcolor=red!25,bordercolor=red
  ]{Xinyu: #1}
}
\nc{\zy}[1]{%
  \todo[linecolor=green!50!black,backgroundcolor=green!25,bordercolor=green!50!black]{Ziyang: #1}%
}
\nc{\yg}[1]{%
  \todo[linecolor=orange!80!black,backgroundcolor=orange!25,bordercolor=orange!80!black]{Yu: #1}%
}
\nc{\ck}[1]{%
  \todo[
    linecolor=violet!80!black,
    backgroundcolor=violet!25,
    bordercolor=violet!80!black
  ]{Chi-Kuang: #1}%
}

\usepackage[normalem]{ulem}

\usepackage{xspace}
\newcommand{\ph}[1]{\underline{\texttt{<#1>}}}

\newcommand{\method}{CSC\xspace}

\title{Severity-Controlled Prediction Sets for Medication Recommendation
}

\author{%
  Yu Gu$^{1}$\thanks{Equal contribution.} \quad
  Zijun Yu$^{1}$\footnotemark[1] \quad
  Chi-Kuang Yeh$^{2}$ \quad
  Xinyu Wang$^{1}$ \quad
  Ziyang Song$^{3}$\thanks{Corresponding author: \texttt{ziyangs@ohio.edu}.} \\
  $^{1}$McGill University, $^{2}$Georgia State University, $^{3}$Ohio University
}

\begin{document}

\maketitle

\begin{abstract}
Medication recommendation is a multi-label prediction task. However, existing prediction-set evaluation and calibration methods rely mainly on exact-overlap metrics and therefore treat all missed medications as equally severe, overlooking how closely the prediction set matches them within the medication hierarchy. To address this, we introduce \textbf{Conformal Severity Control (CSC)}, a post-hoc framework for calibrating medication prediction sets under hierarchy-aware mismatch severity. CSC uses the Anatomical Therapeutic Chemical (ATC) hierarchy to quantify mismatch severity and controls patient-level severity without retraining the predictor, supporting both average severity and upper-quantile severity for the highest-severity mismatches. Experiments on MIMIC-IV show that CSC tracks user-specified severity tolerances and shifts prediction sets away from distant mismatches toward exact or close matches. \blfootnote{Code is available at \url{https://github.com/ggggggyyyyyy/conformal-severity-control}.}

\end{abstract}

\section{Introduction}

\begin{figure}[H]
\centering
\includegraphics[width=0.9\linewidth]{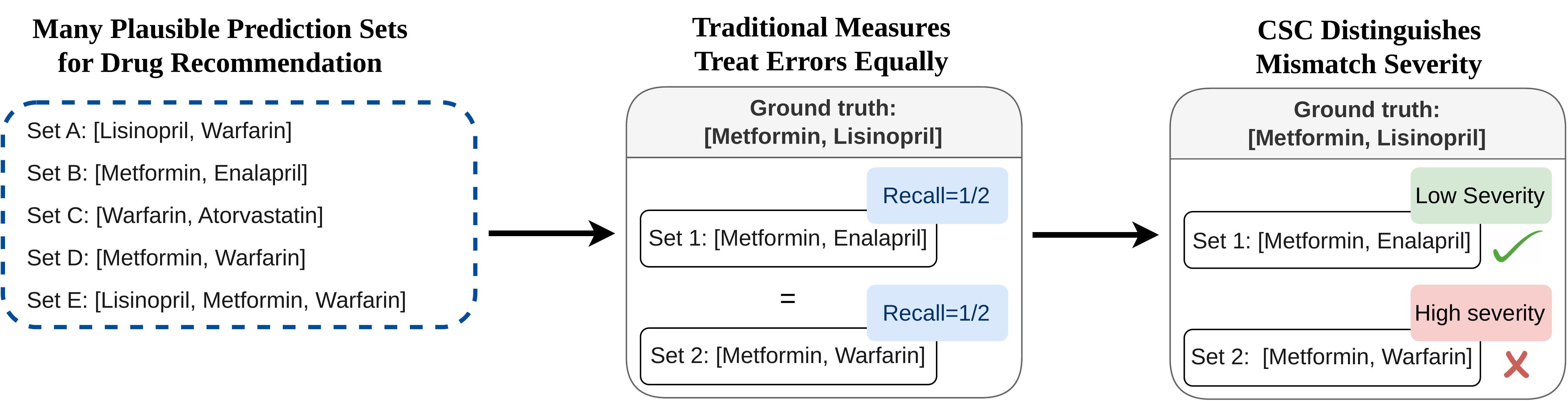}
\vspace{-0.3em}
\caption{\textbf{Motivation for Conformal Severity Control (CSC)}. Prediction sets can achieve the same recall while differing substantially in ATC mismatch severity. CSC selects a score threshold on calibration data so that the expected severity stays below a user-specified tolerance.}\label{fig:csc_motivation}
\vspace{-1em}
\end{figure}

Modern medication recommendation systems increasingly use deep learning models and LLM-based predictors to recommend drugs from electronic health records (EHRs)~\citep{Ali2023-fq, LLM_drug, wang2026saferx}. These models assign scores over large drug vocabularies, which can be used to form candidate medication sets because multiple drugs may be plausible for the same patient~\citep{Kompa2021-vw, begoli2019need}. However, the evaluation and calibration of these sets largely rely on exact-overlap criteria including recall, F1, Jaccard, and false-negative rate~\citep{10.1609/aaai.v33i01.33011126, yang2021safedrug, Ali2023-fq}. These criteria treat every missed medication as the same error, even though a prediction set containing a therapeutically closer medication should incur lower severity than one containing only distant medications. Consequently, two prediction sets can receive the same overlap score while differing substantially in mismatch severity (Figure~\ref{fig:csc_motivation}). This motivates calibrating prediction sets to control hierarchy-aware mismatch severity.

\begin{figure}[t]
    \centering
    \includegraphics[width=0.9\linewidth]{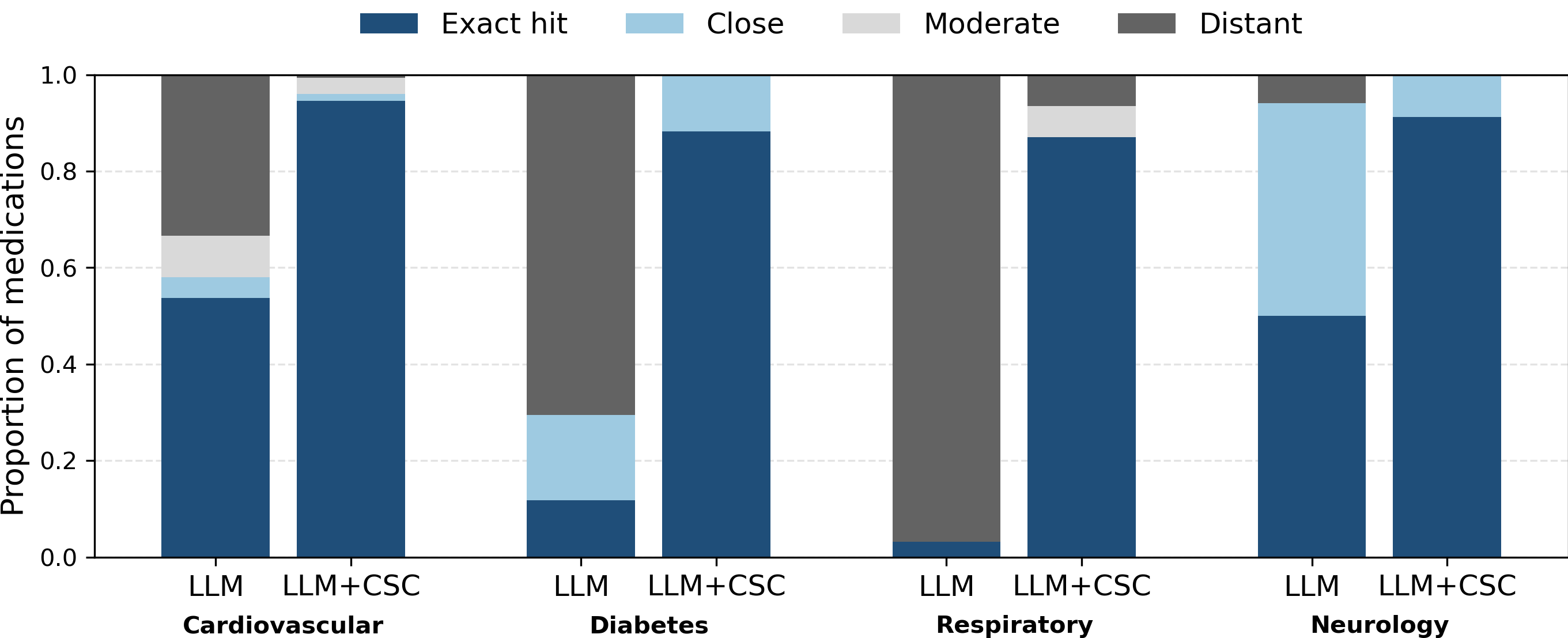}
    \vspace{-0.3em}
    \caption{\textbf{Closest ATC match for disease-associated medications.} For four disease cohorts, each selected ground-truth medication is assigned its closest predicted medication and categorized as exact, close, moderate, or distant. LLM sets threshold the predicted probabilities at 0.25; CSC sets use CSC-Avg with $\alpha=0.125$, whose cutoff is lower. The shift toward exact and close matches therefore reflects set expansion. Details are in Appendix~\ref{app:best_match_details}.}

    \label{fig:best_match_distribution}
    \vspace{-1em}
\end{figure}

To characterize mismatch severity in medication prediction, we analyze LLM-generated prediction sets across cardiovascular, diabetes, respiratory, and neurology cohorts. For each cohort, we focus on selected treatment medications and identify the closest predicted match under the medication hierarchy, categorizing each match as exact, close, moderate, or distant. As shown in Figure~\ref{fig:best_match_distribution}, missed medications in LLM-generated sets are more often matched only to distant medication predictions. In contrast, the CSC-calibrated sets reduce these distant mismatches and increase exact or close matches. This motivates post-hoc calibration under hierarchy-aware mismatch severity.

Risk-control frameworks calibrate a threshold on top of black-box prediction scores so that the expected loss of future prediction sets is bounded~\citep{angelopoulos2023gentle}. Existing frameworks such as Risk-Controlling Prediction Sets (RCPS) and Conformal Risk Control (CRC) provide post-hoc, distribution-free calibration for black-box predictors \citep{bates2021distribution,angelopoulos2024conformal}. However, their guarantees are only as informative as the loss being controlled. When instantiated with overlap-based losses, these frameworks still treat all missed medications equally and therefore cannot distinguish therapeutically closer predictions from distant mismatches. Even after replacing binary overlap with hierarchy-aware mismatch severity, controlling only the average severity can be insufficient. For patients with multiple medications, average severity can obscure a small number of high-severity mismatches when most medications are matched exactly or closely \citep{zhao2024leave}. These limitations call for prediction set calibration that controls mismatch severity and emphasizes the most severe mismatches within each patient.

To address this gap, we introduce \textbf{Conformal Severity Control (CSC)}, a post-hoc, model-agnostic framework for calibrating medication prediction sets under hierarchy-aware mismatch severity. CSC uses the Anatomical Therapeutic Chemical (ATC) hierarchy \citep{who2026atc}  to measure how severe each medication mismatch is and calibrates prediction sets to control a user-specified patient-level severity tolerance, without retraining or modifying the underlying predictor. We further extend CSC from average severity control to upper-quantile severity control~\citep{bental2007old, huang2026optimized}, which emphasizes the highest-severity  mismatches within each patient. On MIMIC-IV, CSC tracks target severity tolerances, induces patient-specific prediction-set sizes under one global threshold, and reduces distant ATC-branch mismatches.

\section{Related Work}

\subsection{Clinical Prediction}

Prior work in clinical prediction has mainly focused on improving predictive models for specific downstream tasks \citep{timelygpt, trajgpt}. In medication recommendation, models have improved candidate drug sets through stronger EHR modeling \citep{Ali2023-fq}, safety-aware objectives \citep{10.1609/aaai.v33i01.33011126}, and richer sources of clinical information \citep{10.1145/3534678.3539089}. A complementary line of work incorporates medical ontologies and structured clinical relationships into the predictive model to improve predictions \citep{Yao2023-ta, 10.1016/j.artmed.2023.102613, Tan2024-ry}. While these approaches improve predictive performance, they are typically specific to individual model architectures.  In contrast, our method is post-hoc and model-agnostic, and can be applied to any predictive model without redesigning the predictor itself.

\subsection{Clinical Calibration via Risk Control}

Conformal prediction and risk-control frameworks provide distribution-free procedures for selecting prediction sets whose expected risk is controlled under exchangeability~\citep{bates2021distribution, angelopoulos2024conformal}. These frameworks are well suited to clinical prediction because they calibrate prediction sets post hoc from the scores of an already trained predictor, without requiring changes to the predictive model itself~\citep{Vazquez2022-xj}. Existing conformal methods have mainly controlled coverage-based losses, such as miscoverage, false negatives, or false positives~\citep{pmlr-v162-fisch22a}. However, such losses do not distinguish therapeutically closer medication predictions from distant errors. We bridge this gap by quantifying the severity of each medication mismatch and calibrating prediction sets under both average and upper-quantile severity.

\begin{figure}[t]
\centering
\includegraphics[width=\textwidth]{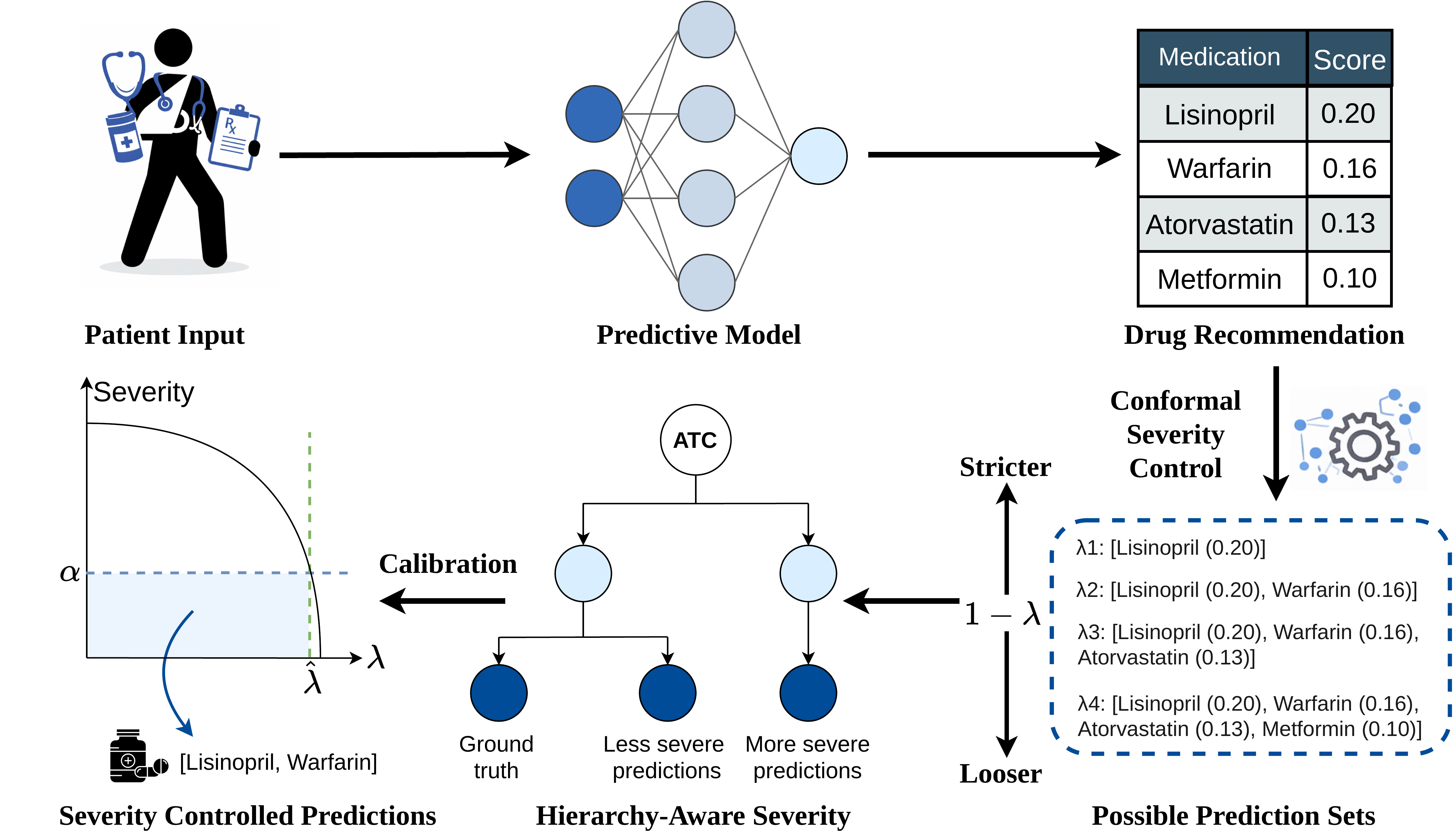}
\vspace{-1em}
\caption{Overview of \method. Given patient input, a predictive model assigns scores to candidate medications, inducing a nested family of prediction sets as the threshold $\lambda$ varies. \method evaluates these candidate sets using an ATC-based hierarchy-aware mismatch severity and calibrates $\lambda$ to control the expected upper-tail patient-level severity.}
\label{fig:csc_overview}
\vspace{-1em}
\end{figure}

\section{Methodology}
\label{sec:methodology}

\subsection{Problem Setup}
\label{sec:problem_setup}

We study multi-label drug recommendation from EHR data, which provide longitudinal patient information for clinical decision support. For each patient, let \(X \in \mathcal{X}\) denote the patient input and let \(Y \subseteq \mathcal{K}\) denote the ground-truth medication set, where \(\mathcal{K}\) is the medication label space. Each medication is represented by an ATC code, which organizes medications hierarchically from broad anatomical and therapeutic groups to finer pharmacological and chemical categories~\citep{who2026atc}. Details of the ATC taxonomy are provided in Appendix~\ref{app:atc_hierarchy}.

As illustrated in Figure~\ref{fig:csc_overview}, a trained predictive model assigns a score $f_k(X) \in [0,1]$ to each medication label $k \in \mathcal{K}$. For a threshold $\lambda \in [0,1]$, we define the induced prediction set as
\begin{align}
C_\lambda(X)
:=
\{k \in \mathcal{K}: f_k(X) \geq 1-\lambda\},
\label{eq:prediction_set}
\end{align}
where $\lambda$ controls the inclusiveness of the prediction set, with smaller values yielding more selective sets and larger values yielding more inclusive sets.

\subsection{Hierarchy-Aware Mismatch Severity}
\label{sec:severity}


Given the prediction set \(C_\lambda(X)\), we quantify how each ground-truth medication is matched by the prediction set. We view the ATC hierarchy as a rooted tree of maximum depth $D$, where medication labels are leaves and internal nodes represent broader medication classes. For a ground-truth medication \(y \in Y\) and a predicted medication \(k \in \mathcal{K}\), let \(\mathrm{LCA}(y,k)\) denote their lowest common ancestor (LCA), i.e., the most specific ATC class containing both medications. If the shared ancestor is deeper in the tree, then \(k\) is closer to \(y\) in the medication taxonomy and should receive lower mismatch severity; if the shared ancestor is closer to the root, then the mismatch is more distant and should receive higher severity.
We define the pairwise mismatch severity between \(y\) and \(k\) as
\begin{align}
s(y,k):= 1-
\frac{\mathrm{depth}\!\left(\mathrm{LCA}(y,k)\right)}{D}.
\label{eq:pairwise_severity}
\end{align}
where \(\mathrm{depth}(\cdot)\) denotes the depth of a node in the ATC tree.
Thus, $s(y,k)=0$ when $y$ and $k$ share the same medication label, while larger values indicate more distant medication classes. Prior work has used ATC distance to quantify therapeutic drug similarity~\citep{wang2013netpredatc} and ATC proximity to characterize therapeutic relatedness~\citep{bohn2025comparators}.

For a ground-truth medication \(y \in Y\), we define its label-level mismatch severity with respect to the prediction set \(C_\lambda(X)\) by the closest predicted medication:
\begin{align}
s\bigl(y,C_\lambda(X)\bigr):=
\begin{cases}
\min_{k\in C_\lambda(X)} s(y,k), & C_\lambda(X)\neq\emptyset,\\
1, & C_\lambda(X)=\emptyset.
\end{cases}
\label{eq:label_severity}
\end{align}
It assigns maximal severity when the prediction set is empty, preserving boundedness and monotonicity. The quantity $s(y,C_\lambda(X))$ is zero when $y\in C_\lambda(X)$, small when the prediction set contains a medication close to \(y\) in the ATC hierarchy, and large when all predicted medications are distant from $y$. Since $C_\lambda(X)$ expands as $\lambda$ increases, $s(y,C_\lambda(X))$ is non-increasing in $\lambda$ for every \(y\in Y\).

\subsection{Average and Upper-Quantile Severity Losses}
\label{sec:severity_loss}

The label-level mismatch severity \(s(y,C_\lambda(X))\) scores a single ground-truth medication. Since a patient may have multiple ground-truth medications, we aggregate label-level severities into a patient-level loss.  For $Y\neq\emptyset$, the average severity loss is
\begin{align}
\mathcal{S}_{\mathrm{avg}}(Y,C_\lambda(X)):=
\frac{1}{|Y|}
\sum_{y\in Y}
s\bigl(y,C_\lambda(X)\bigr),
\label{eq:avg_severity}
\end{align}
and for $Y=\emptyset$ we set this loss, and the upper-quantile loss below, to zero.

Average severity can mask a few high-severity mismatches when most ground-truth medications are matched exactly or closely. To emphasize such mismatches, we control the upper tail of the within-patient severities. For $Y\neq\emptyset$, let $m=|Y|$, write $z_y=s(y,C_\lambda(X))$, and fix a quantile level $\beta\in[0,1)$. The upper-quantile severity loss is the conditional value-at-risk (CVaR) at level $\beta$ of the uniform distribution over $\{z_y\}_{y\in Y}$ \citep{rockafellar2000optimization, bental2007old}:
\begin{align}
\mathcal S_\beta(Y,C_\lambda(X)) := \min_{t\in[0,1]}\Big[t+\frac{1}{(1-\beta)m}\sum_{y\in Y}(z_y-t)_+\Big],
\label{eq:cvar_severity}
\end{align}
where $(u)_+=\max\{u,0\}$. Equivalently, sort $z_{(1)}\ge\cdots\ge z_{(m)}$, and let $q=(1-\beta)m$ and $r=\lceil q\rceil$. Then
\begin{align}
\mathcal S_\beta(Y,C_\lambda(X))=\frac{\sum_{j=1}^{r-1}z_{(j)}+(q-r+1)\,z_{(r)}}{q}.
\label{eq:uq_severity}
\end{align}

The level $\beta$ sets how much of the tail is averaged. At $\beta=0$ the loss is the average severity $\mathcal S_{\mathrm{avg}}$ in~\eqref{eq:avg_severity}, and larger $\beta$ concentrates it on fewer, more severe mismatches, up to the worst label-level severity $\max_{y\in Y} z_y$ once $\beta\ge 1-1/m$. The same $\beta$ thus acts as worst-case severity for patients with few medications and as a tail average for patients with many.



\begin{proposition}[Properties of the severity loss]
\label{prop:severity_properties}
For every $\beta\in[0,1)$, the loss
$\mathcal{S}_{\beta}(Y,C_\lambda(X))$
\begin{enumerate}
  \item[(i)] lies in $[0,1]$;
  \item[(ii)] is non-increasing in $\lambda$;
  \item[(iii)] equals $0$ at $\lambda=1$.
\end{enumerate}
The same holds for $\mathcal S_{\mathrm{avg}}=\mathcal S_0$.
\end{proposition}

These are the conditions of conformal risk control~\citep{angelopoulos2024conformal}: a bounded loss that is non-increasing in $\lambda$ and vanishes at the largest threshold.

\begin{remark}[Connection to FNR control]
\label{remark:conn_fnr}
For $Y\neq\emptyset$,

$$\mathcal S_{\mathrm{avg}}(Y,C_\lambda(X))=\frac{1}{|Y|}\sum_{y\in Y\setminus C_\lambda(X)} s\bigl(y,C_\lambda(X)\bigr)\;\le\;\frac{|Y\setminus C_\lambda(X)|}{|Y|}=\mathrm{FNR}(Y,C_\lambda(X)).$$

CRC-FNR calibrates the loss under the hierarchy-blind cost $s(y,k)=\mathds 1\{k\neq y\}$. At a common $\alpha$ on the same calibration data, CSC-Avg therefore selects $\hat\lambda_{\mathrm{CSC}}\le\hat\lambda_{\mathrm{FNR}}$, giving nested sets $C_{\hat\lambda_{\mathrm{CSC}}}(X)\subseteq C_{\hat\lambda_{\mathrm{FNR}}}(X)$. The gap $\mathrm{FNR}-\mathcal S_{\mathrm{avg}}=\frac{1}{|Y|}\sum_{y\in Y\setminus C_\lambda(X)}\bigl(1-s(y,C_\lambda(X))\bigr)$ is the credit for near misses. CSC thus meets the same tolerance with smaller sets, and the saving comes from missed medications that have an ATC relative in the set.
\end{remark}

\subsection{Conformal Severity Control}
\label{sec:csc_calibration}

We now introduce \method, a post-hoc calibration procedure that selects the threshold $\lambda$ to control patient-level severity. Given a new patient, our goal is to construct prediction sets whose expected patient-level severity is controlled below a user-specified tolerance \(\alpha\). To achieve this without retraining the predictor or assuming a parametric distribution, we work in the CRC framework under exchangeability~\citep{bates2021distribution,angelopoulos2024conformal}. Using the held-out calibration split, we select a threshold $\hat\lambda$ with finite-sample severity control. For a tolerance $\alpha\in(0,1)$, a level $\beta\in[0,1)$, and the loss $\mathcal S_\beta$ from Section~\ref{sec:severity_loss}, the threshold is

\begin{equation}
\hat\lambda
:=
\min\left\{
\lambda\in\Lambda:
\frac{1}{n+1}
\left(
\sum_{i=1}^n
\mathcal{S}_\beta(Y_i,C_\lambda(X_i))
+1
\right)
\leq \alpha
\right\},
\quad
\Lambda=\{j/400: j=0,\ldots,400\},
\label{eq:csc_calibration}
\end{equation}
with $\min\emptyset:=1$. The \(+1\) term accounts for the unseen test patient. Every loss is zero at $\lambda=1$ by Proposition~\ref{prop:severity_properties}(iii), so the feasible set is empty only when $\alpha<1/(n+1)$, and the fallback then returns the full label set. The prediction set for a new patient is $C_{\hat\lambda}(X)$.

\begin{theorem}[Expected severity control]
\label{thm:tail_severity_control}
Let $f$ be a fixed score function, and let $(X_1,Y_1),\dots,(X_n,Y_n),(X,Y)$ be exchangeable. For any $\beta\in[0,1)$ and $\alpha\in(0,1)$, the threshold $\hat\lambda$ in \eqref{eq:csc_calibration} satisfies
\begin{equation}
\mathbb{E}
\left[
\mathcal{S}_\beta(Y,C_{\hat\lambda}(X))
\right]
\leq
\alpha,
\label{eq:tail_severity_guarantee}
\end{equation}
where the expectation is over the calibration data and the new patient.
\end{theorem}

Theorem~\ref{thm:tail_severity_control} is the CRC guarantee~\citep[Theorem~1]{angelopoulos2024conformal} applied to $\mathcal S_\beta$, with Proposition~\ref{prop:severity_properties} supplying its conditions. The guarantee is marginal: it bounds severity averaged over calibration draws and future patients. For $\beta>0$, it controls the expected within-patient tail, rather than an upper quantile of severity across the patient population. Exchangeability is supported by the random patient-level split in Section~\ref{sec: experiment} and does not extend to temporal or cross-site shift.

\subsection{Algorithmic Summary}
\label{sec:algorithmic_summary}

Algorithm~\ref{alg:csc} summarizes \method. Phase 1 runs once on the calibration split. It scans $\Lambda$ in ascending order, computes $\mathcal S_\beta$ for every calibration patient, and stops at the first $\lambda$ that meets the tolerance, returning $\hat\lambda=1$ if none does. Phase 2 applies $\hat\lambda$ to new patients. Theorem~\ref{thm:tail_severity_control} holds for any finite grid containing $\lambda=1$, and refining the grid can only lower $\hat\lambda$; we use a step of $1/400$.

\begin{algorithm}[ht]
\caption{Conformal Severity Control: Offline Calibration and Online Prediction}
\label{alg:csc}
\begin{algorithmic}[1]
\REQUIRE Calibration set $\mathcal{D}_{\text{cal}}=\{(X_i,Y_i)\}_{i=1}^n$, scoring model $f$ trained without $\mathcal{D}_{\text{cal}}$, tolerance $\alpha\in(0,1)$, level $\beta\in[0,1)$, ATC tree of depth $D$, grid $\Lambda=\{j/400:j=0,\ldots,400\}$.
\ENSURE Calibrated threshold $\hat\lambda$ and prediction set $C_{\hat\lambda}(X)$ for a new patient $X$.

\vspace{0.1cm}
\STATE \textbf{\% Phase 1: Offline Calibration}
\STATE $\hat\lambda \gets 1$ \hfill (fallback: full label set)
\FOR{$\lambda\in\Lambda$ in ascending order}
    \FOR{each $(X_i,Y_i)\in\mathcal{D}_{\text{cal}}$}
        \STATE $C_\lambda(X_i)\gets\{k\in\mathcal{K}: f_k(X_i)\geq 1-\lambda\}$
        \IF{$Y_i=\emptyset$}
            \STATE $\mathcal{S}_i(\lambda)\gets 0$
        \ELSE
            \STATE $z_{i,y}\gets s(y,C_\lambda(X_i))$ for each $y\in Y_i$, using $s(y,k)=1-\mathrm{depth}(\mathrm{LCA}(y,k))/D$
            \STATE Sort $z_{i,(1)}\geq\cdots\geq z_{i,(m_i)}$, where $m_i=|Y_i|$; set $q_i=(1-\beta)m_i$, $r_i=\lceil q_i\rceil$
            \STATE $\mathcal{S}_i(\lambda)\gets\bigl(\sum_{j=1}^{r_i-1}z_{i,(j)}+(q_i-r_i+1)\,z_{i,(r_i)}\bigr)/q_i$
        \ENDIF
    \ENDFOR
    \IF{$\frac{1}{n+1}\bigl(\sum_{i=1}^n\mathcal{S}_i(\lambda)+1\bigr)\leq\alpha$}
        \STATE $\hat\lambda\gets\lambda$; \textbf{break}
    \ENDIF
\ENDFOR

\vspace{0.1cm}
\STATE \textbf{\% Phase 2: Online Inference for a New Patient}
\STATE Compute scores $f_k(X)$ for all $k\in\mathcal{K}$
\STATE \textbf{Return} $C_{\hat\lambda}(X)=\{k\in\mathcal{K}: f_k(X)\geq 1-\hat\lambda\}$
\end{algorithmic}
\end{algorithm}

\section{Experiment Design}
\label{sec: experiment}

\textbf{Dataset and Pre-processing.}
We evaluate on MIMIC-IV, a publicly available de-identified EHR dataset of patients admitted to the emergency department or intensive care unit (ICU) at Beth Israel Deaconess Medical Center in Boston, MA \citep{Johnson2023-mimiciv}. It contains inpatient EHR records, including diagnoses, procedures, medication prescriptions,
and admission/discharge timestamps. Diagnoses and procedures are represented using ICD codes, medications are mapped from NDC to ATC level-4 codes. Dataset statistics and preprocessing are detailed in Appendix~\ref{app:dataset}.

\textbf{Medication Recommendation Problem.}
We construct patient sequences by treating each inpatient admission as one visit. For patient \(i\), let \(X_i=[X_i^{(1)},\ldots,X_i^{(N_i)}]\) denote the temporally ordered visit sequence, where each visit contains diagnoses, procedures, and medications. We select the last visit $t = N_i$ of each patient as the target visit. The model input includes patient demographics, diagnoses and procedures from visits \(1\) through \(t\), and medication history from visits \(1\) through \(t-1\). The prediction target is the medication set prescribed at visit \(t\), denoted by \(Y_i^{(t)}\subseteq\mathcal{K}\).

\textbf{LLM Drug Recommender.} We use an LLM-based drug recommendation predictor built on the Llama-based \textit{Meerkat-8B} model~\citep{Meerkat}. The Meerkat-8B LLM takes patient demographics and longitudinal visit history as input using the prompt template provided in Appendix~\ref{app:llm_prompt}. We obtain sequence embeddings using LLM2Vec~\citep{llm2vec} and place a multi-label prediction head on top of these embeddings.  We fine-tune the model with LoRA adapters on the attention projection layers while keeping the remaining parameters frozen~\citep{hu2022lora}. The model is trained with binary cross-entropy (BCE) over medication labels and predicts a score for each medication label. For evaluation, we report precision, recall, and F1.

\textbf{Data Splits and Calibration.} We split patients into training, calibration, and test sets in a 60:20:20 ratio. We use patient-level splits, with all visits from the same patient kept in one split, so exchangeability is assumed between calibration and test patient-level examples. The training split is used to fine-tune the LLM-based predictor for drug recommendation. The calibration split is used to select the CSC threshold \(\hat{\lambda}\), and the test split is held out for final evaluation.


\textbf{Calibration Setup.}
We evaluate CSC under both average severity and upper-quantile severity control at tolerance levels \(\alpha \in \{0.125, 0.25, 0.35\}\).
Since medications are evaluated at ATC level 4, label-level severity takes values in \(\{0,0.25,0.5,0.75,1\}\).  Thus, \(\alpha\) specifies the target average patient-level severity. For example, \(\alpha=0.25\) means that the calibrated prediction sets allow an average severity comparable to a one-level ATC mismatch.
For upper-quantile severity control, we evaluate \(\beta \in \{0.50, 0.75, 0.90\}\), where larger \(\beta\) focuses on fewer high-severity mismatches per patient. The calibrated threshold \(\hat{\lambda}\) is selected on the calibration split and then fixed for evaluation on the test split.

\textbf{Baselines.}
We evaluate two CSC variants, \textbf{CSC-Avg} and \textbf{CSC-Upper}, against both conformal and non-conformal baselines. For the conformal baseline, \textbf{CRC-FNR} calibrates the multi-label false-negative rate (FNR) rather than the hierarchy-aware mismatch severity. For non-conformal baselines, \textbf{Score-threshold} selects all labels whose predicted probability exceeds a global cutoff \(\tau\), while \textbf{Score-top-\(k\)} selects the \(k\) highest-scoring labels for each patient. We tune \(\tau\) or \(k\) on the calibration split to match the recall or prediction set size of CSC. \textbf{Prevalence-top-\(k\)} ignores patient-specific scores and recommends the \(k\) most frequent medication labels in the calibration split.

\textbf{Evaluation Metrics.}
We report empirical average severity, upper-quantile severity, prediction set size, precision, recall, and F1 on the held-out test split. Severity metrics evaluate the calibrated prediction sets under the ATC hierarchy, while precision, recall, and F1 measure exact label overlap.

\section{Results}

\subsection{Severity Calibration Across User-Specified Tolerances}

\begin{figure}[t]
    \centering
    \includegraphics[width=0.9\textwidth]{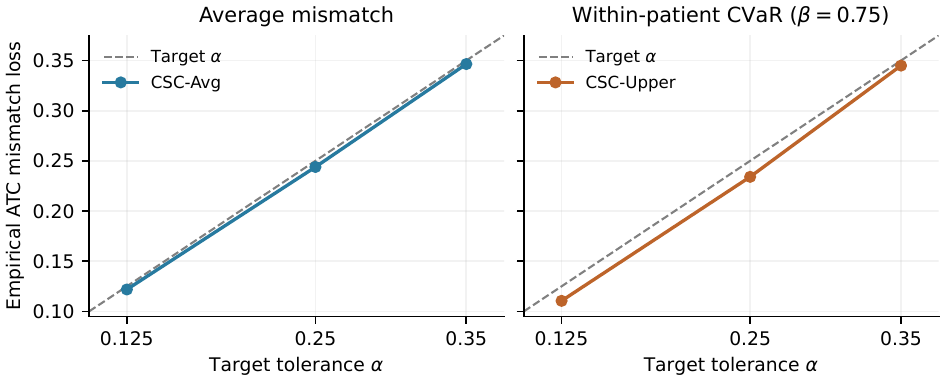}
    \vspace{-0.3em}
    \caption{\textbf{Empirical ATC taxonomy-based mismatch severity across target tolerances.} For each $\alpha$, CSC selects $\hat\lambda$ on the full calibration split, and the resulting prediction sets are evaluated once on the original test split. The left panel shows average severity (CSC-Avg); the right panel shows within-patient empirical CVaR at $\beta=0.75$ (CSC-Upper). The dashed line marks $y=\alpha$. Each point is a single-split empirical estimate.}
    \label{fig:severity_vs_alpha}
    \vspace{-0.7em}
\end{figure}

We evaluate whether \method achieves empirical severity close to the user-specified tolerance \(\alpha\). For each \(\alpha\), we select \(\hat{\lambda}\) on the full calibration split and evaluate the resulting prediction sets once on the original test split. Figure~\ref{fig:severity_vs_alpha} shows that CSC-Avg attains average severity \(0.1218\), \(0.2439\), and \(0.3465\) at \(\alpha=0.125\), \(0.25\), and \(0.35\), and CSC-Upper with \(\beta=0.75\) attains within-patient empirical CVaR \(0.1105\), \(0.2341\), and \(0.3449\). All six point estimates lie below their targets. For a fixed prediction set, larger \(\beta\) defines a stricter loss because it averages fewer and higher-severity mismatches; Table~\ref{tab:severity_by_beta} reports other values of \(\beta\), each calibrated separately.

Table~\ref{tab:alpha_summary} shows the resulting trade-off: for CSC-Avg, increasing \(\alpha\) from \(0.125\) to \(0.35\) reduces the mean set size from \(28.53\) to \(11.66\), raising precision from \(0.315\) to \(0.522\) and lowering recall from \(0.813\) to \(0.551\). At the same \(\alpha\), CSC-Upper returns larger sets because it controls the worst-covered medications within each patient.

\begin{figure}[t]
    \centering    \includegraphics[width=\textwidth]{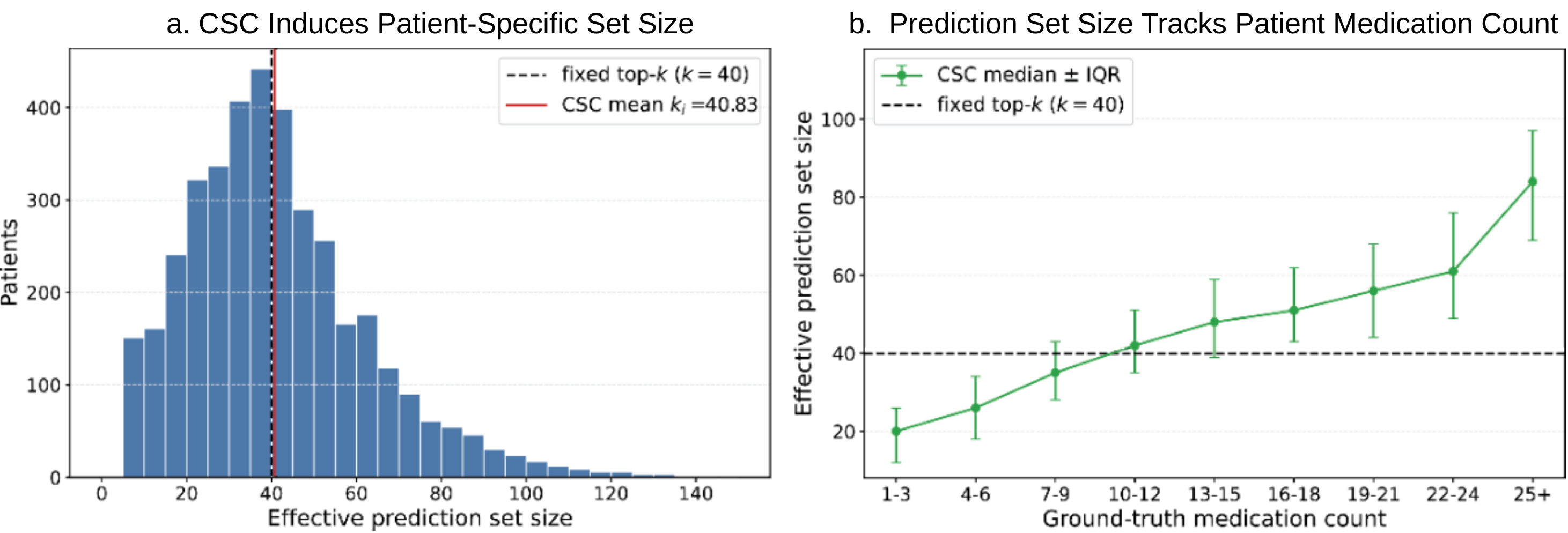}
    \vspace{-1em}
    \caption{\textbf{CSC produces patient-specific prediction sets under a single calibrated threshold \(\hat{\lambda}\).} (a) CSC matches fixed Top-\(k\) baseline in average set size but produces a broad distribution of \(k_i\) across patients. (b) Larger ground-truth medication sets correspond to larger calibrated sets, while fixed Top-\(k\) returns the same set size for all patients without considering patient-specific mismatch severity.}
    \label{fig:dynamic_topk}
\end{figure}

\begin{table}[t]
\centering
\caption{\textbf{Calibration results across target tolerances.} For each $\alpha$, $\hat\lambda$ is selected on the full calibration split, and all other entries are point estimates on the original test split. The controlled loss is average severity for CSC-Avg and within-patient empirical CVaR at $\beta=0.75$ for CSC-Upper. Precision, recall, and F1 are averaged across patients.}
\label{tab:alpha_summary}
\footnotesize
\setlength{\tabcolsep}{5pt}
\begin{tabular}{lccccccc}
\toprule
Method & $\alpha$ & $\hat{\lambda}$ & Set Size & Controlled Loss & Precision & Recall & F1 \\
\midrule
\multirow{3}{*}{CSC-Avg}
& 0.125 & 0.9325 & 28.53 & 0.1218 & 0.315 & 0.813 & 0.438 \\
& 0.25  & 0.8225 & 16.38 & 0.2439 & 0.443 & 0.661 & 0.510 \\
& 0.35  & 0.7175 & 11.66 & 0.3465 & 0.522 & 0.551 & 0.512 \\
\midrule
\multirow{3}{*}{CSC-Upper $(\beta=0.75)$}
& 0.125 & 0.9875 & 62.87 & 0.1105 & 0.166 & 0.945 & 0.274 \\
& 0.25  & 0.9675 & 40.83 & 0.2341 & 0.240 & 0.888 & 0.366 \\
& 0.35  & 0.9425 & 30.94 & 0.3449 & 0.298 & 0.833 & 0.423 \\
\bottomrule
\end{tabular}
\vspace{-2em}
\end{table}

\textbf{Finite-sample correction.} To isolate the $+1$ correction in Eq.~\eqref{eq:csc_calibration}, we compare CSC-Avg with a plug-in rule that uses the same scores, loss, and grid but selects the smallest $\lambda$ whose uncorrected calibration loss is at most $\alpha$. At $\alpha=0.25$, we draw 100 calibration subsamples without replacement for each $n\in\{25,50,100\}$ and evaluate both rules on the same fixed test split. Table~\ref{tab:cr_correction} shows that the plug-in test loss exceeds $\alpha$ in 38--41\% of subsamples, whereas CSC exceeds it in 13--28\%. CSC pays for this with larger sets, 2.07 additional labels on average at $n=25$. With the full calibration split, the two rules nearly coincide. These exceedance fractions are empirical diagnostics on a fixed test split, not failure probabilities of Theorem~\ref{thm:tail_severity_control}, which bounds the expected loss. Appendix~\ref{app:cr_correction} extends the comparison to $n\le 2{,}000$.

\begin{table}[t]
\centering
\caption{\textbf{Effect of the finite-sample correction at $\alpha=0.25$ (CSC-Avg).} Plug-in omits the $+1$ correction in Eq.~\eqref{eq:csc_calibration}. For $n\le 100$, entries are means over 100 calibration subsamples evaluated on the same fixed test split; brackets give empirical 2.5--97.5\% quantiles of test loss across subsamples, and Exceed is the fraction of subsamples with test loss above $\alpha$. The full-calibration row is a single point estimate.}
\label{tab:cr_correction}
\small
\begin{tabular}{rcccccc}
\toprule
& \multicolumn{3}{c}{CSC} & \multicolumn{3}{c}{Plug-in} \\
\cmidrule(lr){2-4}\cmidrule(lr){5-7}
$n$ & Loss [quantiles] & Size & Exceed & Loss [quantiles] & Size & Exceed \\
\midrule
25    & 0.214 [0.154, 0.287] & 18.67 & 13\% & 0.245 [0.171, 0.319] & 16.60 & 40\% \\
50    & 0.230 [0.183, 0.284] & 17.44 & 19\% & 0.245 [0.198, 0.303] & 16.45 & 41\% \\
100   & 0.238 [0.203, 0.275] & 16.80 & 28\% & 0.246 [0.210, 0.287] & 16.35 & 38\% \\
3,826 & 0.244 & 16.38 & -- & 0.246 & 16.23 & -- \\
\bottomrule
\end{tabular}
\end{table}

\subsection{Patient-Adaptive Set Sizes under a Global Severity Threshold}

We next examine the set sizes that a single global threshold induces. For patient \(i\), the threshold \(\hat{\lambda}\) yields an effective set size \(k_i=|C_{\hat{\lambda}}(X_i)|\) that depends on the patient's score distribution, whereas fixed Top-\(k\) assigns the same \(k\) to every patient. Figure~\ref{fig:dynamic_topk}a shows CSC-Upper at \(\alpha=0.25\) and \(\beta=0.75\): its mean set size is close to that of size-matched Top-\(k\) (\(\bar{k}_i=40.83\) versus \(k=40\)), but \(k_i\) varies widely across patients (median \(38\), interquartile range \([26,52]\)). Figure~\ref{fig:dynamic_topk}b shows that \(k_i\) increases with the ground-truth medication count \(|Y_i|\). This variation describes how the calibrated cutoff distributes set sizes across patients; it does not by itself imply lower ATC mismatch at a matched budget, which Table~\ref{tab:severity_radius_coverage} examines.

\begin{table}[t]
\centering
\caption{\textbf{Matched baseline diagnostics at $\alpha=0.25$.} All rules use the same predictor scores. Matching parameters are selected on the calibration split, and all entries are point estimates on the original test split. CRC-FNR matched to CSC's calibration recall or mean set size selects the same cutoff as CSC and returns identical sets. Score-top-$k$ is matched to CSC's mean calibration set size. Bold marks the loss that each CSC variant controls.}
\label{tab:severity_radius_coverage}
\footnotesize
\setlength{\tabcolsep}{4pt}
\begin{tabular}{lcccccc}
\toprule
Method & Set size & $\widehat S_{\mathrm{avg}}$ & $\widehat S_{0.75}$ & Precision & Recall & F1 \\
\midrule
CSC-Avg (= matched CRC-FNR) & 16.38 & \textbf{0.2439} & 0.6247 & 0.443 & 0.661 & 0.510 \\
Score-top-$k$ ($k=16$)      & 16.00 & 0.2206 & 0.6032 & 0.414 & 0.677 & 0.462 \\
\midrule
CSC-Upper (= matched CRC-FNR) & 40.83 & 0.0680 & \textbf{0.2341} & 0.240 & 0.888 & 0.366 \\
Score-top-$k$ ($k=40$)        & 40.00 & 0.0649 & 0.2361 & 0.235 & 0.889 & 0.346 \\
\bottomrule
\end{tabular}
\end{table}

\subsection{Calibration and Set-Size Trade-offs under Matched Baselines}

Table~\ref{tab:severity_radius_coverage} compares CSC with baselines that use the same predictor scores at $\alpha=0.25$. All matching parameters are selected on the calibration split before test evaluation. CRC-FNR uses the same nested threshold family as CSC; when its target is matched to CSC's calibration recall or mean set size, it selects the same cutoff and returns identical sets. Score-top-$k$ uses a different set-construction rule. At matched mean size, it attains lower average loss than CSC-Avg ($0.2206$ at $k=16$ versus $0.2439$) and similar within-patient CVaR to CSC-Upper ($0.2361$ at $k=40$ versus $0.2341$), while CSC attains higher precision and F1 in both comparisons. CSC therefore does not yield lower ATC mismatch at a fixed budget. Its role is to choose the cutoff directly from a prespecified ATC-risk tolerance with the finite-sample correction in Eq.~\eqref{eq:csc_calibration}; Table~\ref{tab:cr_correction} isolates the effect of that correction. Appendix~\ref{app:matched_baselines} reports all tolerances and the prevalence baseline.

\section{Discussion and Limitations}
\label{sec:discussion}
CSC controls an ATC taxonomy-based mismatch cost. The consistency between the mismatch cost and the practical clinical harms need to be further studied. Our evidence covers one single-center dataset, MIMIC-IV, its 300 most frequent ATC-4 labels, and two predictors, an LLM-based one and a non-LLM one (Appendix~\ref{app:cr_drugdoctor}), and the marginal guarantee assumes exchangeability. Future work includes clinician-grounded evaluation and extending CSC to temporal and cross-hospital distribution shift.

\section{Conclusion}

We introduced CSC, a post-hoc framework for calibrating medication prediction sets under ATC-based mismatch severity.
CSC replaces binary missed-label errors with a hierarchy-aware severity loss and calibrates a single threshold to control expected patient-level severity for any trained scoring model under exchangeability.
On MIMIC-IV, CSC converts user-specified severity tolerances into calibrated prediction sets: \(\alpha\) sets the allowed severity level, while \(\beta\) controls emphasis on the highest-severity mismatches.
The calibrated sets track the target severity levels, adapt their size across patients under one global threshold.


\clearpage

\begin{ack}
Computing resources were provided via an NVIDIA Academic Grant Program award to Ziyang Song at Ohio University.
Zijun Yu acknowledges partial support from an ISM Graduate Scholarship. Chi-Kuang Yeh acknowledges partial support from start-up funds provided by Georgia State University.
\end{ack}

\bibliographystyle{unsrtnat}
\bibliography{refs}

\clearpage

\appendix

\section{Notations of Variables}

\begin{table}[H]
\centering
\caption{Summary of notation used throughout the paper.}
\label{tab:notation}
\small
\begin{tabular}{>{\raggedright\arraybackslash}p{0.27\textwidth} >{\raggedright\arraybackslash}p{0.66\textwidth}}
\toprule
\textbf{Notation} & \textbf{Description} \\
\midrule

\multicolumn{2}{l}{\textbf{Spaces and label sets}} \\
\midrule
$\mathcal{X}$ & Input space. \\
$\mathcal{K}$ & Medication label space. \\
$\mathcal{Y}$ & Output space of label sets, e.g., subsets of $\mathcal{K}$. \\
$k \in \mathcal{K}$ & Candidate or predicted medication label. \\
$Y \subseteq \mathcal{K}$ & Ground-truth medication label set for a patient. \\
$y \in Y$ & Single ground-truth medication label. \\

\midrule
\multicolumn{2}{l}{\textbf{Random variables, observations, and calibration data}} \\
\midrule
$X \in \mathcal{X}$ & Patient/model input, viewed as a random variable. \\
$\{(X_i, Y_i)\}_{i=1}^n$ & Exchangeable calibration dataset, viewed as random input--label pairs. \\
$x_i$ & Observed features of visit $i$ in a patient's longitudinal record. \\
$n$ & Number of calibration examples. \\
$i$ & Calibration-example index. \\
$\mathbb{E}[\cdot]$ & Expectation with respect to future exchangeable data. \\

\midrule
\multicolumn{2}{l}{\textbf{Prediction sets and thresholding}} \\
\midrule
$f_k(X)$ & Model score assigned to label $k$ for input $X$. \\
$\lambda \in [0,1]$ & Threshold parameter controlling the prediction set size. \\
$C_\lambda(X)$ & Prediction set at threshold $\lambda$. \\
$\lambda_1, \lambda_2$ & Thresholds used to express the nestedness of prediction sets. \\
$\hat\lambda$ & Calibrated threshold for the generic or upper-quantile severity loss. \\
$\hat\lambda_{\mathrm{avg}}$ & Calibrated threshold for average-severity control. \\

\midrule
\multicolumn{2}{l}{\textbf{Hierarchy-aware severity}} \\
\midrule
$s : \mathcal{K} \times \mathcal{K} \to [0,1]$ & Hierarchy-aware pairwise severity function. \\
$s(y,k)$ & Severity of predicting label $k$ when the ground-truth label is $y$. \\
$\mathrm{LCA}(y,k)$ & Lowest common ancestor of $y$ and $k$ in the ATC hierarchy. \\
$\mathrm{depth}(\cdot)$ & Depth of a node in the ATC hierarchy. \\
$s(y, C_\lambda(X))$ & Severity of the closest label in $C_\lambda(X)$ to the true label $y$. \\

\midrule
\multicolumn{2}{l}{\textbf{Patient-level losses and conformal calibration}} \\
\midrule
$\ell(C_\lambda(X), Y)$ & Generic bounded monotone loss used in the conformal calibration framework. \\
$\mathcal{L}$ & Generic patient-level severity loss used by CSC. \\
$\mathcal{L}_i(\lambda)$ & Patient-level loss of calibration example $i$ at threshold $\lambda$. \\
$\beta \in [0,1)$ & Severity quantile level \\
$\mathcal{S}_{\beta}(Y, C_\lambda(X))$ & Patient-level upper-tail severity loss at severity quantile level $\beta$. \\
$\mathcal{S}_{\mathrm{avg}}(Y, C_\lambda(X))$ & Patient-level average severity loss. This is a special case of $\mathcal{S}_{\beta}(Y, C_\lambda(X))$ where $\beta=0$.\\
$\alpha \in (0,1)$ & User-specified target tolerance for the population loss or severity. \\
\midrule

\multicolumn{2}{l}{\textbf{Drug recommendation task notation}} \\
\midrule
$[X_i^{(1)},\ldots,X_i^{(N_i)}]$ & Temporally ordered patient sequence. \\
$Y_i^{(t)}$ & Target medications set at visit $t$. \\
\bottomrule
\end{tabular}
\end{table}

\clearpage

\section{Proofs of the Propositions and Main Theorem}


\subsection{Proof of \cref{prop:severity_properties}}
\begin{proof}[Proof of \cref{prop:severity_properties}]
Part (i).
For nonempty $Y$, define
\[
F_\beta(t,\lambda)
:=t+\frac{1}{(1-\beta)|Y|}
\sum_{y\in Y}\bigl(s(y,C_\lambda(X))-t\bigr)_+.
\]
Since $F_\beta(t,\lambda)\geq0$ for $t\in[0,1]$ and $s(y,C_\lambda(X))\leq1$,
\[
0\leq\mathcal{S}_\beta(Y,C_\lambda(X))
=\min_{t\in[0,1]}F_\beta(t,\lambda)
\leq F_\beta(1,\lambda)=1.
\]
For empty $Y$, the loss is zero by convention.

Part (ii).
For fixed $Y\ne\emptyset$ and $\beta$, each label-level severity is non-increasing in $\lambda$. Thus, for $\lambda_1\leq\lambda_2$,
\[
F_\beta(t,\lambda_2)\leq F_\beta(t,\lambda_1)
\quad\text{for every }t\in[0,1].
\]
Taking minima over the same interval gives
\[
\begin{aligned}
\mathcal{S}_\beta(Y,C_{\lambda_2}(X))
&=\min_{t\in[0,1]}F_\beta(t,\lambda_2)\\
&\leq\min_{t\in[0,1]}F_\beta(t,\lambda_1)
=\mathcal{S}_\beta(Y,C_{\lambda_1}(X)).
\end{aligned}
\]
For empty $Y$, the loss remains zero.

Part (iii) holds because $C_1(X)=\{k:f_k(X)\geq 0\}=\mathcal{K}\supseteq Y$, so $s(y,C_1(X))=s(y,y)=0$ for every $y\in Y$, and the loss of an all-zero collection is $0$.
\end{proof}

\subsection{Proof of \cref{thm:tail_severity_control}}

\begin{proof}[Proof of \cref{thm:tail_severity_control}]

We establish the result by verifying that $\mathcal{S}_\beta(Y, C_\lambda(X))$ satisfies the conditions of conformal risk
control~\citep{angelopoulos2024conformal}, then applying that framework carefully
to handle the selection of $\hat{\lambda}$.

\paragraph{Step 1: Loss properties.}
By Proposition~\ref{prop:severity_properties}, the severity satisfies two
properties required by the conformal risk control framework.
\begin{enumerate}
    \item \textbf{Boundedness.}
          $\mathcal{S}_\beta(y, C_\lambda(x)) \in [0,1]$ for all
          $(x,y) \in \mathcal{X}\times\mathcal{Y}$ and all $\lambda\in\Lambda$.
    \item \textbf{Monotonicity.}
          $\mathcal{S}_\beta(y, C_{\lambda_1}(x)) \geq
           \mathcal{S}_\beta(y, C_{\lambda_2}(x))$ whenever $\lambda_1 \leq \lambda_2$,
          since larger $\lambda$ induces larger prediction sets,
          $C_{\lambda_1}(x) \subseteq C_{\lambda_2}(x)$, and severity is
          non-increasing in set size.
\end{enumerate}

\paragraph{Step 2: Calibration on a finite grid.}
The threshold $\hat\lambda$ in~\eqref{eq:csc_calibration} is selected from the finite grid $\Lambda=\{j/400:j=0,\ldots,400\}$, which contains $\lambda=1$, with the fallback $\hat\lambda=1$ when no $\lambda\in\Lambda$ satisfies the calibration condition. The argument below uses only that $\Lambda$ is finite and contains $1$. The threshold $\hat\lambda$ is a function of the calibration data $\{(X_i,Y_i)\}_{i=1}^n$ only.

\paragraph{Step 3: Expected loss on the test point.}
Define the augmented empirical risk
\[
\widehat{R}(\lambda;\mathcal{D}_{n+1})
:=
\frac{1}{n+1}
\sum_{i=1}^{n+1}
\mathcal{S}_\beta(Y_i, C_\lambda(X_i)),
\]
where
\[
\mathcal{D}_{n+1}
=
\{(X_i,Y_i)\}_{i=1}^{n+1},
\qquad
(X_{n+1},Y_{n+1}) = (X,Y).
\]

Consider the auxiliary threshold
\[
\lambda^*
=
\min\!\left\{
\lambda \in \Lambda :
\widehat{R}(\lambda;\mathcal{D}_{n+1})
\le \alpha
\right\}.
\]
By Proposition~\ref{prop:severity_properties}(iii), $\widehat{R}(1;\mathcal{D}_{n+1})=0\le\alpha$, so this set is nonempty and $\lambda^*$ is well defined.

Since $\lambda^*$ is a symmetric function of the augmented dataset
$\mathcal{D}_{n+1}$, and the samples
$\{(X_i,Y_i)\}_{i=1}^{n+1}$ are jointly exchangeable,
the random variables
\[
\mathcal{S}_\beta(Y_i,C_{\lambda^*}(X_i)),
\qquad i=1,\dots,n+1,
\]
are exchangeable. Therefore,
\begin{align}
\mathbb{E}\!\left[
\mathcal{S}_\beta(Y,C_{\lambda^*}(X))
\right]
&=
\frac{1}{n+1}
\sum_{i=1}^{n+1}
\mathbb{E}\!\left[
\mathcal{S}_\beta(Y_i,C_{\lambda^*}(X_i))
\right]
\notag\\
&=
\mathbb{E}\!\left[
\widehat{R}(\lambda^*;\mathcal{D}_{n+1})
\right]
\le \alpha.
\label{eq:lambda_star_bound}
\end{align}

It remains to compare $\lambda^*$ with the calibrated threshold
$\hat\lambda$ defined in~\eqref{eq:csc_calibration}.
By Step~1,
\[
0 \le
\mathcal{S}_\beta(Y,C_\lambda(X))
\le 1
\qquad \text{for all } \lambda.
\]
Hence, for every $\lambda \in \Lambda$,
\[
\widehat{R}(\lambda;\mathcal{D}_{n+1})
=
\frac{1}{n+1}
\sum_{i=1}^{n+1}
\mathcal{S}_\beta(Y_i,C_\lambda(X_i))
\le
\frac{1}{n+1}
\left(
\sum_{i=1}^n
\mathcal{S}_\beta(Y_i,C_\lambda(X_i))
+1
\right).
\]

Therefore, every $\lambda$ satisfying the calibration condition in
\eqref{eq:csc_calibration} is also feasible for the definition of
$\lambda^*$, which implies
\[
\lambda^* \le \hat\lambda.
\]
If no $\lambda\in\Lambda$ satisfies the calibration condition, then $\hat\lambda=1\geq\lambda^*$, so the inequality holds in both cases.

Finally, by monotonicity of
$\mathcal{S}_\beta(Y,C_\lambda(X))$ with respect to $\lambda$,
\[
\mathcal{S}_\beta(Y,C_{\hat\lambda}(X))
\le
\mathcal{S}_\beta(Y,C_{\lambda^*}(X))
\qquad \text{almost surely}.
\]
Taking expectations and applying~\eqref{eq:lambda_star_bound} gives
\[
\mathbb{E}\!\left[
\mathcal{S}_\beta(Y,C_{\hat\lambda}(X))
\right]
\le
\mathbb{E}\!\left[
\mathcal{S}_\beta(Y,C_{\lambda^*}(X))
\right]
\le \alpha,
\]
which proves~\eqref{eq:tail_severity_guarantee}.

\paragraph{Step 4: The $\beta = 0$ case.}
When $\beta = 0$, the severity functional reduces to the average severity,
$\mathcal{S}_0 = \mathcal{S}_{\mathrm{avg}}$. The boundedness and monotonicity
properties in Step~1 hold for $\mathcal{S}_{\mathrm{avg}}$ by
Proposition~\ref{prop:severity_properties}, so Steps~2--3 apply verbatim,
yielding~\eqref{eq:tail_severity_guarantee} for all $\beta \in [0,1)$.
\end{proof}

\section{Dataset Details}
\label{app:dataset}
\subsection{ATC Hierarchy and Severity Interpretation}
\label{app:atc_hierarchy}

The World Health Organization's Anatomical Therapeutic Chemical (ATC) system organizes medications into a five-level hierarchy, from broad anatomical categories to specific chemical substances \citep{who2026atc}. In our setting, this hierarchy contains 14 anatomical main groups, 90 therapeutic subgroups, 248 pharmacological subgroups, 840 chemical subgroups, and 5,495 chemical substances. Each ATC code can therefore be viewed as a path in a rooted tree, where longer shared prefixes indicate greater therapeutic similarity~\citep{song2026smi, li2026hcoe}. For example, \texttt{A10BA02} denotes metformin, and its successive prefixes \texttt{A}, \texttt{A10}, \texttt{A10B}, and \texttt{A10BA} encode its ancestor categories in the ATC hierarchy. In our experiments, medications are represented at \textbf{ATC level-4}, as described in Appendix~\ref{app:preprocess}.

This ATC hierarchy gives a direct interpretation of the severity function in Eq.~\eqref{eq:pairwise_severity}. For ATC-4 labels, a lowest common ancestor at depth 4 (same ATC-4 code) yields severity $0$; depth 3 (same pharmacological subgroup) yields severity $\nicefrac{1}{4}$; depth 2 (same therapeutic class) yields severity $\nicefrac{1}{2}$; depth 1 (same anatomical group) yields severity $\nicefrac{3}{4}$; and depth 0 (different top-level groups) yields severity $1$. Thus, prediction sets are penalized less for including medications closer to the ground-truth in the ATC hierarchy and more for including only distant medications.

\subsection{Dataset and Cohort Statistics}

We used a large-scale, de-identified MIMIC-IV EHR dataset from Beth Israel Deaconess Medical Center (BIDMC), released via PhysioNet under a credentialed data use agreement \citep{Johnson2023-mimiciv}. This dataset contains longitudinal inpatient records, including diagnoses, procedures, medication prescriptions,
and admission/discharge timestamps. Patient identifiers are removed and all timestamps are shifted by a patient-specific offset. This preserves relative temporal ordering within each patient sequence while preventing alignment of absolute calendar time across patients. We construct visit-based patient sequences by treating each inpatient admission as a visit and ordering visits
chronologically by admission time. Table~\ref{tab:mimic_basic_stats} reports basic cohort statistics after preprocessing.

\begin{table}[t]
\centering
\caption{Structure of the ATC medication hierarchy used in this study. \emph{\# Concepts} denotes the number of unique medication concepts at each level.}
\label{tab:atc_hierarchy}
\footnotesize
\begin{tabular}{lcccc}
\toprule
Level & Meaning & \# Concepts & Example Code & Example Code Description \\
\midrule
1 & Anatomical main group & 14 & \texttt{A} & Alimentary tract and metabolism \\
2 & Therapeutic main group & 90 & \texttt{A10} & Drugs used in diabetes \\
3 & Pharmacological subgroup & 248 & \texttt{A10B} & Blood glucose lowering drugs, excluding insulins \\
4 & Chemical subgroup & 840 & \texttt{A10BA} & Biguanides \\
5 & Chemical substance & 5,495 & \texttt{A10BA02} & Metformin \\
\bottomrule
\end{tabular}
\end{table}

\begin{table}[th]
\centering
\caption{Basic statistics of the pre-processed MIMIC-IV EHR dataset. ``\#'' denotes counts and ``/patient'' denotes the average number of records per-patient.}
\label{tab:mimic_basic_stats}
\small
\setlength{\tabcolsep}{5pt}
\begin{tabular}{lcccccc}
\toprule
Dataset
& \#patients
& \#visits
& \#visits/patient
& \#conditions/patient
& \#procedures/patient
& \#drugs/patient \\
\midrule
MIMIC-IV
& 19,135 & 35,974 & 1.88 & 21.74 & 4.70 & 43.89 \\
\bottomrule
\end{tabular}
\end{table}
\subsection{Preprocessing and Concept Normalization}
\label{app:preprocess}

We applied the same preprocessing pipeline to the MIMIC-IV dataset, including selection criteria, code normalization, and data splitting, yielding consistent visit-based patient EHR sequences. We applied the same normalization procedure across cohorts to ensure a fair comparison.

\paragraph{Cohort construction and splitting.}
We treat each inpatient admission as a visit and represent each patient as a chronologically ordered
sequence of visits.
We perform patient-level training/calibration/test splits so that all visits from the same patient appear in only one split, avoiding information leakage across splits.

\paragraph{Diagnosis and procedure codes.}
For each visit, we extract diagnoses and procedures from the corresponding MIMIC-IV tables and represent them
as unordered sets of codes.
We retain the original ICD coding systems (ICD-9-CM and ICD-10-CM for diagnoses; ICD-9 and ICD-10-PCS for
procedures) and preserve the recorded ICD version to avoid conflating codes across systems. We map ICD-9/10 diagnosis and procedure codes to CCS categories, and de-duplicate repeated codes within each visit. Specifically, 11,736 ICD-9-CM and 72,446 ICD-10-CM diagnosis codes are mapped to 285 CCS condition categories, while 4,670 ICD-9-PROC and 79,758 ICD-10-PCS procedure codes are mapped to 231 CCS procedure categories, yielding 285 condition and 231 procedure concepts in total.

\paragraph{Medication mapping and normalization.}
We follow the ATC-L4 medication normalization setting used in SafeRx-Agent~\citep{wang2026saferx}. Medications are extracted from the prescription tables, where drugs are recorded as National Drug Codes (NDCs). We standardize them using a two-stage mapping pipeline:
\textbf{NDC $\rightarrow$ RxNorm (RxCUI) $\rightarrow$ ATC}.
We retain \textbf{ATC level-4} codes to preserve meaningful therapeutic and chemical structure. Unmappable NDC entries (e.g., non-drug supplies, institution-specific packaging codes, or missing references) are excluded from the medication representation. Finally, we retain the 300 most frequent ATC level-4 medication codes as the prediction vocabulary. The severity loss is evaluated over this retained medication vocabulary.

\paragraph{Visit-level feature construction.}
Each visit $x_i$ is represented by three unordered sets: diagnosis codes, procedure codes, and mapped ATC-4 medication codes.
These per-visit sets are then composed into a time-ordered patient sequence $\{x_1,\ldots,x_t\}$ by admission time.
Basic cohort statistics after preprocessing are reported in Table~\ref{tab:mimic_basic_stats}.

\subsection{LLM Prompt Template}
\label{app:llm_prompt}

\begin{tcolorbox}[
  enhanced,
  colback=white,
  colframe=gray!60,
  boxrule=0.9pt,
  arc=3mm,
  left=4mm,right=4mm,top=5mm,bottom=4mm,
  width=\linewidth,
  before upper={\raggedright},
  title=Prompt Template for Drug Recommendation,
  colbacktitle=gray!70,
  coltitle=white,
  fonttitle=\Large\rmfamily,
  boxed title style={boxrule=0pt,arc=3mm,left=4mm,right=4mm,top=2mm,bottom=2mm},
]
\noindent \textbf{Demographics:} Insurance: \ph{INSURANCE}; Language: \ph{LANGUAGE};\\
Religion: \ph{RELIGION}; Marital status: \ph{MARITAL\_STATUS}; Ethnicity: \ph{ETHNICITY}. \par
\noindent The patient has \ph{VISIT\_NUM} ICU visits. \par
\textcolor{blue}{\noindent In visit 1, diagnoses: \ph{DIAG\_NAME}, \ldots; procedures: \ph{PROC\_NAME}, \ldots; medications: \ph{MED\_NAME}, \ldots.}\par
\textcolor{blue}{\noindent \ldots}\par
\noindent Current visit: diagnoses: \ph{DIAG\_NAME}, \ldots; procedures: \ph{PROC\_NAME}, \ldots. \\
\noindent \textbf{Task:} Predict the medications to prescribe for the current visit. Output a multi-label set of \textbf{4-level ATC codes}. Choose any number of codes from the medication vocabulary. Note that a patient may have multiple medications.\\
\noindent \textbf{Output format:} A multi-label set of ATC level-4 codes.
\end{tcolorbox}

\subsection{Computing Resources}
\label{app:compute_resources}
Experiments for training the LLM predictor are conducted on an NVIDIA DGX Spark system with 128 GB of memory. Training each LLM takes 2.45 hours per epoch. For MIMIC-IV, we further split the training portion into training and validation subsets. The model is fine-tuned for up to $10$ epochs, with early stopping based on validation performance.
Mixed-precision training (FP16) is used to improve training efficiency.

\section{Additional Results}

\subsection{LLM and CSC Prediction Set Matching Analysis}
\label{app:best_match_details}

We provide additional implementation details and result discussion for the prediction set matching analysis shown in Figure~\ref{fig:best_match_distribution}. The goal of this analysis is to characterize whether treatment-associated medications are exactly matched, or instead matched by therapeutically close or distant medications in the prediction set. Beyond exact-overlap metrics, this analysis measures the hierarchical distance between each ground-truth medication and its closest predicted medication under the ATC taxonomy. To achieve this, we construct four disease cohorts representing distinct clinical categories and identify a set of representative treatment medications for each disease category. We compare the prediction sets generated by the baseline LLM drug recommender with the CSC-calibrated prediction sets constructed from the same LLM predictor. For each ground-truth treatment medication, we identify the closest predicted medication in the ATC hierarchy and categorize the match as exact, close, moderate, or distant according to the shared hierarchy depth.

\paragraph{Cohort Construction.}
We analyze four disease cohorts: cardiovascular disease, diabetes, respiratory disease, and neurology.
The cardiovascular cohort contains 346 patients and corresponds to hypertension or chronic cardiovascular disease.
The diabetes cohort contains 11 patients and corresponds to type 2 diabetes or metabolic syndrome.
The respiratory cohort contains 29 patients and corresponds to COPD or asthma.
The neurology cohort contains 25 patients and corresponds to epilepsy or chronic neurologic disease.
These cohorts are analyzed independently, so the same patient may appear in more than one cohort because comorbidities are common.

For each cohort, we evaluate only ground-truth medications that belong to ATC branches associated with the target disease category, not all medications in the patient EHR data.
For cardiovascular disease, we use C02, C03, C07, C08, and C09.
For diabetes, we use A10.
For respiratory disease, we use R03.
For neurology, we use N03A.
Across the four cohorts, each patient has an average of 2.91 cardiovascular-associated, 1.55 diabetes-associated, 1.07 respiratory-associated, and 1.36 neurology-associated ground-truth medication labels, respectively.

\paragraph{Prediction Sets from the LLM Drug Recommender and CSC.}
We use the LLM-based drug predictor described in Section~\ref{sec: experiment} as the base prediction model and apply the CSC framework to its medication prediction scores. This yields two prediction sets for comparison. The first is the LLM prediction set obtained by thresholding the LLM-predicted medication probabilities at 0.25. The second is the CSC-calibrated prediction set obtained from the same LLM prediction scores using CSC-Avg with target severity tolerance \(\alpha=0.125\). For each selected target medication, we compare it with all medications in the prediction set and select the one with the closest ATC hierarchy match. We then categorize the best match as exact, close, moderate, or distant according to its ATC hierarchy-aware mismatch severity: exact hits have severity \(0\), close non-exact matches have severity in \((0,0.25]\), moderate matches have severity \(0.5\), and distant mismatches have severity at least \(0.75\).

\paragraph{Overall Results Across Disease Cohorts.}
Across all evaluated disease-associated medication labels from the four cohorts, CSC achieves better best-match outcomes than the LLM prediction sets.
The exact-hit rate increases from 51.56\% for the LLM prediction sets to 94.13\% for the CSC-calibrated prediction sets.
The exact-or-close match rate increases from 57.16\% to 95.96\%.
The distant mismatch rate decreases from 34.86\% to 0.83\%, and the mean best-match severity decreases from 0.328 to 0.0268.
These results show that CSC shifts prediction sets away from distant mismatches and toward exact or close medication matches.

\paragraph{Macro-Averaged Results.}
Because the cardiovascular cohort contains many more labels than the other cohorts, we also compute macro-averaged results across the four disease groups.
The macro-average exact-hit rate increases from 29.69\% for the LLM prediction sets to 90.26\% for the CSC-calibrated prediction sets.
The macro-average distant mismatch rate decreases from 51.64\% to 1.79\%, and the mean best-match severity decreases from 0.490 to 0.0394.
These results show that the improvement is not driven only by the large cardiovascular cohort; CSC consistently shifts disease-associated medication matches away from distant mismatches and toward exact hits across disease areas.

\paragraph{Cardiovascular Cohort Results.}
In the cardiovascular cohort, the exact-hit rate increases from 53.77\% for the LLM prediction sets to 94.54\% for the CSC-calibrated prediction sets.
The distant mismatch rate decreases from 33.33\% to 0.69\%.
In absolute counts, the LLM prediction sets produce 542 exact hits, 43 close non-exact matches, 87 moderate matches, and 336 distant mismatches.
The CSC-calibrated prediction sets produce 953 exact hits, 15 close non-exact matches, 33 moderate matches, and only 7 distant mismatches.
These results show that CSC substantially increases exact hits and nearly removes distant ATC-branch mismatches in the largest cohort.

\paragraph{Diabetes Cohort Results.}
In the diabetes cohort, the exact-hit rate increases from 11.76\% for the LLM prediction sets to 88.24\% for the CSC-calibrated prediction sets.
The distant mismatch rate decreases from 70.59\% to 0\%.
In absolute counts, the LLM prediction sets produce 2 exact hits, 3 close non-exact matches, and 12 distant mismatches.
The CSC-calibrated prediction sets produce 15 exact hits, 2 close non-exact matches, and no distant mismatches.
Although this cohort is small, these results show that CSC substantially increases exact hits and removes distant ATC-branch mismatches.

\paragraph{Respiratory Cohort Results.}
In the respiratory cohort, the exact-hit rate increases from 3.23\% for the LLM prediction sets to 87.10\% for the CSC-calibrated prediction sets.
The distant mismatch rate decreases from 96.77\% to 6.45\%.
In absolute counts, the LLM prediction sets produce 1 exact hit and 30 distant mismatches.
The CSC-calibrated prediction sets produce 27 exact hits, 2 moderate matches, and 2 distant mismatches.
These results show that CSC produces the largest reduction in distant ATC-branch mismatches in the respiratory cohort.

\paragraph{Neurology Cohort Results.}
In the neurology cohort, the exact-hit rate increases from 50.00\% for the LLM prediction sets to 91.18\% for the CSC-calibrated prediction sets.
The distant mismatch rate decreases from 5.88\% to 0\%.
In absolute counts, the LLM prediction sets produce 17 exact hits, 15 close non-exact matches, and 2 distant mismatches.
The CSC-calibrated prediction sets produce 31 exact hits, 3 close non-exact matches, and no distant mismatches.
These results show that CSC increases exact hits and removes the remaining distant ATC-branch mismatches in this cohort.

\paragraph{Transitions from Distant Mismatches to Exact Hits or Close Matches.}
We further measure how often the same ground-truth medication changes from a distant mismatch under the LLM prediction sets to an exact hit or close match under the CSC-calibrated prediction sets.
This transition occurs for 30.65\% of cardiovascular labels, 70.59\% of diabetes labels, 83.87\% of respiratory labels, and 5.88\% of neurology labels.
The mean severity reduction is largest in respiratory disease, where it reaches 0.831.
The corresponding reductions are 0.544 for diabetes, 0.286 for cardiovascular disease, and 0.140 for neurology.
These results show that CSC not only increases exact hits, but also moves many high-severity distant mismatches into exact or low-severity categories.

\paragraph{Key Findings.}
This prediction-set matching analysis shows why exact-overlap evaluation is not sufficient for medication recommendation.
Exact-overlap metrics measure whether a ground-truth medication is exactly included in the prediction set; however, they do not distinguish a therapeutically close medication match in the ATC hierarchy from a distant match.
The proposed hierarchy-aware mismatch severity captures this difference by measuring how far the closest predicted medication is from each ground-truth medication.
The disease-cohort results show that the LLM prediction sets often leave disease-associated medications matched only to distant ATC branches, especially in the diabetes and respiratory cohorts.
In contrast, the CSC-calibrated prediction sets substantially reduce distant mismatches and shift the best-match distribution toward exact hits and close matches.
These findings show that CSC improves the structure of prediction errors, not only the number of exact hits.

\subsection{Effect of Upper-Quantile Level \(\beta\)}

Increasing $\beta$ has two effects. For a fixed prediction set, it moves the loss toward the most severe mismatches within each patient; under separate calibration, it also changes the selected threshold. Table~\ref{tab:severity_by_beta} separates the two at $\alpha=0.25$. Calibration with $\beta=0.75$ selects $\hat\lambda=0.9675$. Holding this threshold fixed gives mean set size $40.83$ for every $\beta$, and the patient-averaged empirical CVaR rises from $0.1335$ at $\beta=0.50$ to $0.2341$ at $\beta=0.75$ and $0.3521$ at $\beta=0.90$. The last value exceeds $0.25$ because this threshold was calibrated for $\beta=0.75$. Calibrating each $\beta$ separately instead selects larger thresholds as $\beta$ grows, with mean set sizes $27.48$, $40.83$, and $54.08$ and losses $0.2463$, $0.2341$, and $0.2364$, all below the target. Larger $\beta$ thus gives stronger protection against the worst-covered medications at the cost of larger sets.

\begin{table}[t]
\centering
\caption{\textbf{Effect of $\beta$ at $\alpha=0.25$: fixed threshold versus separate calibration.} The fixed-threshold column keeps the CSC-Upper threshold calibrated at $\beta=0.75$ ($\hat\lambda=0.9675$, mean set size 40.83) and changes only $\beta$ in the loss. The separate-calibration columns select a new threshold for each $\beta$. Losses are patient-averaged within-patient empirical CVaR; all entries are point estimates on the original test split.}
\label{tab:severity_by_beta}
\small
\setlength{\tabcolsep}{6pt}
\begin{tabular}{ccccc}
\toprule
& Fixed threshold & \multicolumn{3}{c}{Separate calibration} \\
\cmidrule(lr){2-2}\cmidrule(lr){3-5}
$\beta$ & Test loss & $\hat\lambda$ & Test loss & Set size \\
\midrule
0.50 & 0.1335 & 0.9275 & 0.2463 & 27.48 \\
0.75 & 0.2341 & 0.9675 & 0.2341 & 40.83 \\
0.90 & 0.3521 & 0.9825 & 0.2364 & 54.08 \\
\bottomrule
\end{tabular}
\end{table}

\subsection{Finite-Sample Correction across Calibration Sizes}
\label{app:cr_correction}

This appendix extends Table~\ref{tab:cr_correction} to larger calibration sizes. Both rules use the same predictor scores, loss, and grid $\Lambda$; CSC applies the $+1$ correction in Eq.~\eqref{eq:csc_calibration}, and the plug-in rule omits it. For each $n$, we draw 100 calibration subsamples without replacement from the 3,826 calibration patients. Within a repetition, the subsamples for different $n$ are nested prefixes of one random permutation, and both rules use the same subsample. All rules are evaluated on the same fixed test split. As $n$ grows, the correction shrinks and the two rules converge. For $n\ge 1{,}000$, subsamples drawn from the same 3,826 patients overlap substantially, so their spread understates the variability of independent calibration sets.

\begin{table}[h]
\centering
\caption{\textbf{Finite-sample correction across calibration sizes at $\alpha=0.25$ (CSC-Avg).} For $n\le 2{,}000$, loss and size are means over 100 calibration subsamples on a fixed test split; brackets give empirical 2.5--97.5\% quantiles of test loss, and Exceed is the fraction of subsamples with test loss above $\alpha$. Exceed is a fixed-test empirical diagnostic, not a failure probability of Theorem~\ref{thm:tail_severity_control}. The row $n=3{,}826$ uses the full calibration split once.}
\label{tab:cr_correction_full}
\small
\begin{tabular}{rcccccc}
\toprule
& \multicolumn{3}{c}{CSC} & \multicolumn{3}{c}{Plug-in} \\
\cmidrule(lr){2-4}\cmidrule(lr){5-7}
$n$ & Loss [quantiles] & Size & Exceed & Loss [quantiles] & Size & Exceed \\
\midrule
25    & 0.214 [0.154, 0.287] & 18.67 & 13\% & 0.245 [0.171, 0.319] & 16.60 & 40\% \\
50    & 0.230 [0.183, 0.284] & 17.44 & 19\% & 0.245 [0.198, 0.303] & 16.45 & 41\% \\
100   & 0.238 [0.203, 0.275] & 16.80 & 28\% & 0.246 [0.210, 0.287] & 16.35 & 38\% \\
250   & 0.242 [0.220, 0.262] & 16.53 & 23\% & 0.245 [0.224, 0.266] & 16.36 & 30\% \\
500   & 0.244 [0.229, 0.259] & 16.36 & 20\% & 0.246 [0.229, 0.261] & 16.28 & 26\% \\
1,000 & 0.245 [0.236, 0.254] & 16.33 & 13\% & 0.246 [0.236, 0.254] & 16.29 & 18\% \\
2,000 & 0.245 [0.239, 0.249] & 16.31 & 2\%  & 0.245 [0.240, 0.249] & 16.29 & 2\%  \\
3,826 & 0.244 & 16.38 & -- & 0.246 & 16.23 & -- \\
\bottomrule
\end{tabular}
\end{table}

\subsection{ATC Hierarchy Branch Statistics}
\label{app:branch_stats}

We further inspect how CSC allocates prediction mass across the ATC hierarchy.
For each ATC level-2 branch, we aggregate medication-level true positives and false positives among predicted medications.
We also report branch-level recall and purity, where purity is the branch-level precision
\(\mathrm{TP}/(\mathrm{TP}+\mathrm{FP})\).
These plots are diagnostic only.

\begin{figure}[t]
\centering
\includegraphics[width=0.85\linewidth]{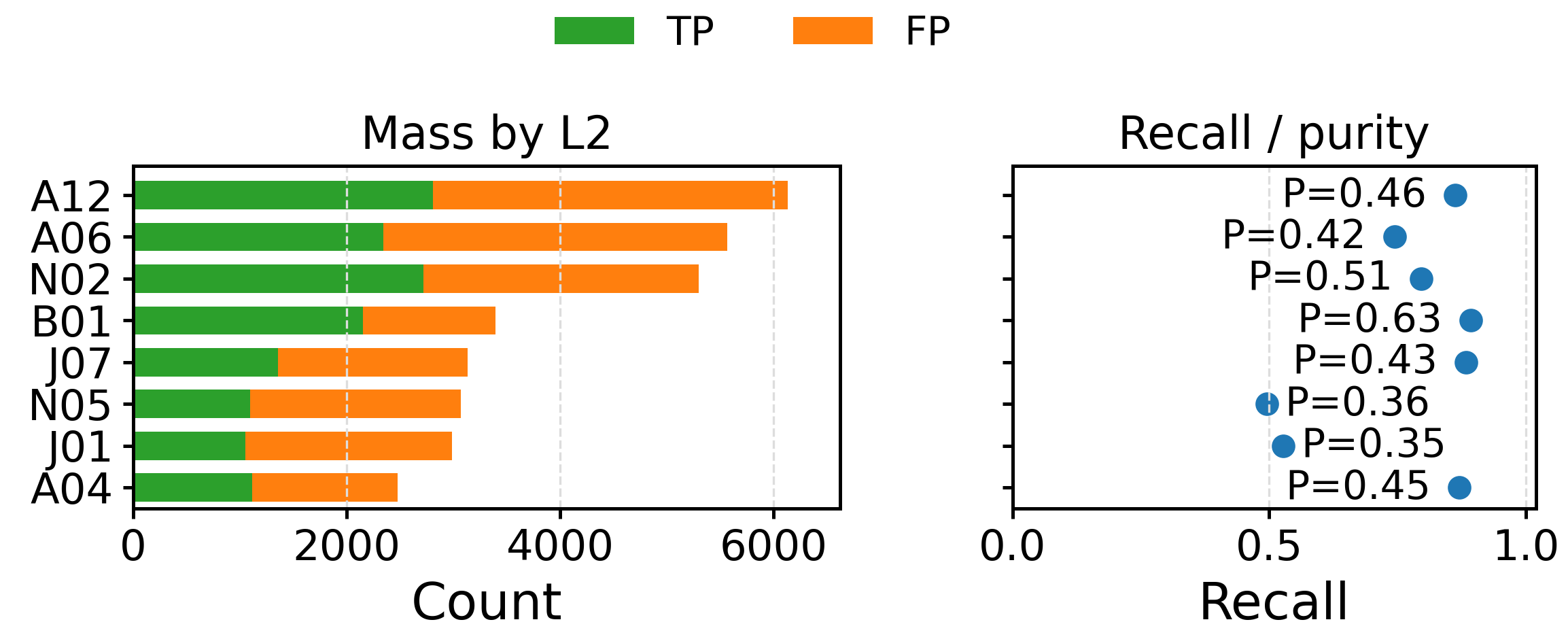}
\caption{ATC level-2 branch composition for average-severity CSC at $\alpha=0.25$. The left panel shows true positives and false positives among the most frequently selected ATC branches, while the right panel reports branch-level recall and precision.}
\label{fig:avg_branch_stats_alpha025_l2}
\end{figure}

Figure~\ref{fig:avg_branch_stats_alpha025_l2} shows the branch-level composition for CSC-Avg at \(\alpha=0.25\).
Most prediction mass is concentrated in common ATC level-2 branches, including A12 mineral supplements, A06 drugs for constipation, N02 analgesics, B01 antithrombotic agents, and J07 vaccines.
CSC-Avg keeps prediction sets relatively compact, so branch-level purity remains moderate in several common branches, but recall varies across medication families.

\begin{figure}[ht]
\centering
\includegraphics[width=0.85\linewidth]{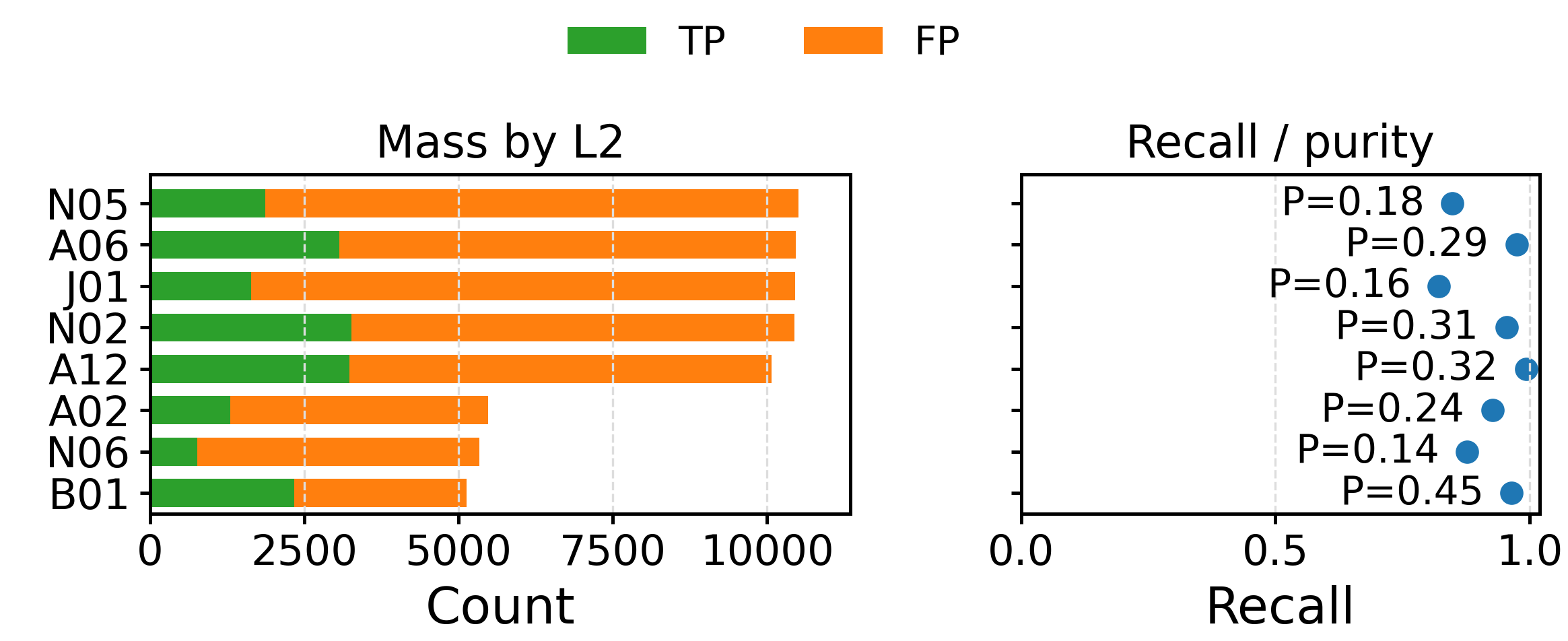}
\caption{ATC level-2 branch composition for upper-quantile severity CSC at $\alpha=0.25$ and $\beta=0.75$. Compared with average-severity calibration, upper-tail calibration selects larger sets and increases branch-level recall in many medication families, at the cost of lower branch-level purity.}
\label{fig:upper_branch_stats_alpha025_l2}
\end{figure}

Figure~\ref{fig:upper_branch_stats_alpha025_l2} shows the corresponding branch-level composition for CSC-Upper with \(\alpha=0.25\) and \(\beta=0.75\).
Because CSC-Upper controls the worst-covered medications within each patient, it selects larger prediction sets and expands coverage across more ATC branches.
This increases branch-level recall in many branches, but also introduces more false positives and lowers purity.
The pattern is consistent with the intended trade-off of upper-tail severity control: reducing high-severity misses requires more conservative prediction sets.

\subsection{Matched Baseline Comparisons}
\label{app:matched_baselines}

Table~\ref{tab:size_matched} extends Table~\ref{tab:severity_radius_coverage} to all three tolerances. All matching parameters are selected on the calibration split and applied once to the original test split, so the test set sizes of matched baselines can differ slightly from those of CSC. Size matching selects the $k$ whose mean calibration set size is closest to that of CSC, and recall matching does the same for patient-averaged recall; ties go to the more inclusive candidate. Score-top-$k$ ranks medications by predictor score for each patient, with ties broken by label index. Prevalence-top-$k$ returns the $k$ medications most frequent in the calibration split to every patient.

CRC-FNR is not listed separately: in all six settings, matching its target to CSC's calibration recall or mean set size selects CSC's cutoff and returns identical sets. Score-threshold uses the same threshold family, and at CSC's cutoff it reproduces CSC's sets. Score-top-$k$ attains lower average loss than CSC-Avg at every tolerance under both matching criteria. Its within-patient CVaR differs from that of CSC-Upper by at most $0.012$, lower at one setting and higher at the others. CSC attains higher patient-averaged precision and F1 than Score-top-$k$ in every comparison. These diagnostics therefore show no fixed-budget advantage for CSC in ATC mismatch; they report the set-size and exact-overlap cost of the operating point that CSC selects for a prespecified tolerance. Prevalence-top-$k$, which ignores patient-specific scores, has much higher loss at matched size and needs much larger sets to match recall.

\begin{table}[t]
\centering
\caption{\textbf{Matched baseline diagnostics at all tolerances.} Matching parameters are selected on the calibration split, and all entries are point estimates on the original test split. CRC-FNR matched to CSC's calibration recall or mean set size returns the same sets as CSC in every block. ``Size'' and ``recall'' mark the matching criterion. Median [IQR] summarizes set sizes across test patients; fixed-$k$ rules give every patient the same size. Bold marks the loss that CSC controls in each block.}
\label{tab:size_matched}
\footnotesize
\setlength{\tabcolsep}{3pt}
\begin{tabular}{lccccccc}
\toprule
Rule & Mean size & Median [IQR] & $\widehat S_{\mathrm{avg}}$ & $\widehat S_{0.75}$ & Precision & Recall & F1 \\
\midrule
\multicolumn{8}{l}{\textit{CSC-Avg, $\alpha=0.125$ ($\hat\lambda=0.9325$)}} \\
CSC-Avg (= matched CRC-FNR)            & 28.53 & 27 [17, 37] & \textbf{0.1218} & 0.3833 & 0.315 & 0.813 & 0.438 \\
Score-top-$k$ (size, $k=28$)           & 28.00 & --          & 0.1108 & 0.3754 & 0.303 & 0.821 & 0.405 \\
Score-top-$k$ (recall, $k=27$)         & 27.00 & --          & 0.1171 & 0.3917 & 0.310 & 0.813 & 0.410 \\
Prevalence-top-$k$ (size, $k=28$)      & 28.00 & --          & 0.2146 & 0.6128 & 0.246 & 0.655 & 0.327 \\
Prevalence-top-$k$ (recall, $k=50$)    & 50.00 & --          & 0.1043 & 0.3529 & 0.175 & 0.817 & 0.270 \\
\midrule
\multicolumn{8}{l}{\textit{CSC-Avg, $\alpha=0.25$ ($\hat\lambda=0.8225$)}} \\
CSC-Avg (= matched CRC-FNR)            & 16.38 & 14 [8, 23]  & \textbf{0.2439} & 0.6247 & 0.443 & 0.661 & 0.510 \\
Score-top-$k$ (size, $k=16$)           & 16.00 & --          & 0.2206 & 0.6032 & 0.414 & 0.677 & 0.462 \\
Score-top-$k$ (recall, $k=15$)         & 15.00 & --          & 0.2360 & 0.6262 & 0.426 & 0.658 & 0.464 \\
Prevalence-top-$k$ (size, $k=16$)      & 16.00 & --          & 0.3700 & 0.7778 & 0.314 & 0.504 & 0.347 \\
Prevalence-top-$k$ (recall, $k=28$)    & 28.00 & --          & 0.2146 & 0.6128 & 0.246 & 0.655 & 0.327 \\
\midrule
\multicolumn{8}{l}{\textit{CSC-Avg, $\alpha=0.35$ ($\hat\lambda=0.7175$)}} \\
CSC-Avg (= matched CRC-FNR)            & 11.66 & 9 [5, 17]   & \textbf{0.3465} & 0.7573 & 0.522 & 0.551 & 0.512 \\
Score-top-$k$ (size, $k=12$)           & 12.00 & --          & 0.2913 & 0.6965 & 0.467 & 0.593 & 0.466 \\
Score-top-$k$ (recall, $k=10$)         & 10.00 & --          & 0.3388 & 0.7445 & 0.501 & 0.541 & 0.463 \\
Prevalence-top-$k$ (size, $k=12$)      & 12.00 & --          & 0.4455 & 0.8341 & 0.341 & 0.426 & 0.339 \\
Prevalence-top-$k$ (recall, $k=19$)    & 19.00 & --          & 0.3098 & 0.7164 & 0.296 & 0.555 & 0.348 \\
\midrule
\multicolumn{8}{l}{\textit{CSC-Upper ($\beta=0.75$), $\alpha=0.125$ ($\hat\lambda=0.9875$)}} \\
CSC-Upper (= matched CRC-FNR)          & 62.87 & 58 [41, 80] & 0.0307 & \textbf{0.1105} & 0.166 & 0.945 & 0.274 \\
Score-top-$k$ (size, recall; $k=62$)   & 62.00 & --          & 0.0289 & 0.1121 & 0.164 & 0.945 & 0.264 \\
Prevalence-top-$k$ (size, $k=62$)      & 62.00 & --          & 0.0744 & 0.2642 & 0.149 & 0.865 & 0.241 \\
Prevalence-top-$k$ (recall, $k=105$)   & 105.00 & --         & 0.0246 & 0.0912 & 0.098 & 0.944 & 0.171 \\
\midrule
\multicolumn{8}{l}{\textit{CSC-Upper ($\beta=0.75$), $\alpha=0.25$ ($\hat\lambda=0.9675$)}} \\
CSC-Upper (= matched CRC-FNR)          & 40.83 & 38 [26, 52] & 0.0680 & \textbf{0.2341} & 0.240 & 0.888 & 0.366 \\
Score-top-$k$ (size, recall; $k=40$)   & 40.00 & --          & 0.0649 & 0.2361 & 0.235 & 0.889 & 0.346 \\
Prevalence-top-$k$ (size, $k=40$)      & 40.00 & --          & 0.1488 & 0.4776 & 0.202 & 0.758 & 0.296 \\
Prevalence-top-$k$ (recall, $k=70$)    & 70.00 & --          & 0.0622 & 0.2243 & 0.136 & 0.887 & 0.224 \\
\midrule
\multicolumn{8}{l}{\textit{CSC-Upper ($\beta=0.75$), $\alpha=0.35$ ($\hat\lambda=0.9425$)}} \\
CSC-Upper (= matched CRC-FNR)          & 30.94 & 29 [19, 40] & 0.1073 & \textbf{0.3449} & 0.298 & 0.833 & 0.423 \\
Score-top-$k$ (size, $k=31$)           & 31.00 & --          & 0.0960 & 0.3330 & 0.283 & 0.843 & 0.390 \\
Score-top-$k$ (recall, $k=30$)         & 30.00 & --          & 0.1010 & 0.3475 & 0.289 & 0.836 & 0.395 \\
Prevalence-top-$k$ (size, $k=31$)      & 31.00 & --          & 0.1946 & 0.5794 & 0.233 & 0.690 & 0.320 \\
Prevalence-top-$k$ (recall, $k=55$)    & 55.00 & --          & 0.0966 & 0.3302 & 0.163 & 0.834 & 0.257 \\
\bottomrule
\end{tabular}
\end{table}

\subsection{Application to DrugDoctor-Based Scores}
\label{app:cr_drugdoctor}

We apply CSC unchanged to scores from a non-LLM recommender based on DrugDoctor~\citep{kuang2024drugdoctor}. The scores were supplied as precomputed matrices for the same calibration and test patients. We verified that patient IDs and label matrices match those of the LLM predictor, and the authors confirmed the ordering of the 300-label vocabulary; only the scores differ. The loss matrix, grid, and calibration rule are unchanged. Table~\ref{tab:cr_drugdoctor} shows that all six test losses lie below their targets, and CSC-Upper again selects larger sets with higher recall and lower precision. The supplied files do not document whether DrugDoctor's training and model selection excluded the calibration and test patients, so we report these results as an empirical application and do not invoke Theorem~\ref{thm:tail_severity_control} for them.

\begin{table}[t]
\centering
\caption{\textbf{CSC with DrugDoctor-based scores.} Each row calibrates on all 3,826 calibration patients and reports point estimates on the 3,827 test patients. Bold marks the loss that each CSC variant controls.}
\label{tab:cr_drugdoctor}
\footnotesize
\setlength{\tabcolsep}{4pt}
\begin{tabular}{lcccccccc}
\toprule
Method & $\alpha$ & $\hat\lambda$ & $\widehat S_{\mathrm{avg}}$ & $\widehat S_{0.75}$ & Set size & Precision & Recall & F1 \\
\midrule
CSC-Avg & 0.125 & 0.9375 & \textbf{0.1197} & 0.3708 & 43.86 & 0.204 & 0.816 & 0.316 \\
CSC-Avg & 0.25  & 0.8575 & \textbf{0.2464} & 0.6091 & 24.73 & 0.309 & 0.659 & 0.399 \\
CSC-Avg & 0.35  & 0.7700 & \textbf{0.3462} & 0.7316 & 16.73 & 0.385 & 0.541 & 0.419 \\
\midrule
CSC-Upper & 0.125 & 0.9825 & 0.0312 & \textbf{0.1103} & 93.81 & 0.105 & 0.940 & 0.185 \\
CSC-Upper & 0.25  & 0.9600 & 0.0740 & \textbf{0.2436} & 58.86 & 0.159 & 0.876 & 0.262 \\
CSC-Upper & 0.35  & 0.9450 & 0.1054 & \textbf{0.3337} & 47.63 & 0.191 & 0.834 & 0.301 \\
\bottomrule
\end{tabular}
\end{table}

\subsection{Sensitivity to the Taxonomy Cost}
\label{app:cr_cost_sensitivity}

We recalibrate CSC-Avg at $\alpha=0.25$ under three monotone transformations of the normalized LCA distance $d=s(y,k)$ in Eq.~\eqref{eq:pairwise_severity}, namely $\sqrt d$, $d$, and $d^2$, which change the penalty for nonzero mismatches but preserve its ordering. We also use a branch-dependent cost built from the ATC tree. For a node $v$, let $N(v)$ be its number of ATC-5 descendants and $I(v)=-\log\{N(v)/N(\mathrm{root})\}$; the cost between labels $u$ and $v$ is $I(u)+I(v)-2I(\mathrm{LCA}(u,v))$, normalized by its maximum over the 300-label vocabulary. Among the 302 label pairs whose LCA has depth 3, the equal-edge cost is $0.25$ for every pair, whereas this cost ranges from $0.081$ to $0.449$. Every cost is bounded and zero for exact matches, so the loss stays monotone and the calibration procedure is unchanged. For label \texttt{L01XC}, which is absent from the hierarchy file, we keep its code-prefix path and count one pseudo-leaf.

Table~\ref{tab:cr_cost_sensitivity} shows that every cost stays below its own target, while the cutoff, set size, and recall change with the cost. Losses on different scales are compared only with the target in their own row. These costs remain taxonomy-based and do not measure clinical harm.

\begin{table}[t]
\centering
\caption{\textbf{Taxonomy-cost sensitivity for CSC-Avg at $\alpha=0.25$.} Each cost is calibrated separately on the full calibration split, and entries are point estimates on the original test split. Each test loss is on its own scale and is compared only with the $0.25$ target.}
\label{tab:cr_cost_sensitivity}
\small
\setlength{\tabcolsep}{5pt}
\begin{tabular}{lcccccc}
\toprule
Cost & $\hat\lambda$ & Test loss & Set size & Precision & Recall & F1 \\
\midrule
$\sqrt d$            & 0.8550 & 0.2464 & 18.64 & 0.412 & 0.700 & 0.499 \\
$d$ (equal-edge)     & 0.8225 & 0.2439 & 16.38 & 0.443 & 0.661 & 0.510 \\
$d^2$                & 0.7700 & 0.2440 & 13.68 & 0.485 & 0.603 & 0.515 \\
Branch-dependent IC  & 0.6950 & 0.2477 & 10.94 & 0.537 & 0.529 & 0.508 \\
\bottomrule
\end{tabular}
\end{table}

\end{document}